%% file: main.tex
\documentclass[journal]{IEEEtran}
\usepackage[utf8]{inputenc}
\usepackage[main=english]{babel}
\usepackage[T1]{fontenc}
\usepackage{csquotes}
\usepackage{xspace}
\usepackage{times}
\usepackage{latexsym}
\usepackage{setspace}
\usepackage{microtype}
\usepackage{graphicx}
\usepackage{dsfont}
\usepackage{caption}
\usepackage{nicematrix}
\usepackage{xcolor}
\usepackage[breaklinks]{hyperref}
\usepackage{amsfonts}
\usepackage{amsmath}
\usepackage{amssymb}
\usepackage{stackrel}
\usepackage{tikz}
\usepackage{booktabs}
\usepackage{multirow}
\usepackage{adjustbox}
\usepackage[breakable,skins]{tcolorbox}

\ifCLASSOPTIONcompsoc
\usepackage[nocompress]{cite}
\else
\usepackage{cite}
\fi

\title{Semi-supervised Multimodal Representation Learning through a Global Workspace}
\author{Benjamin~Devillers, L\'eopold~Mayti\'e \& Rufin~VanRullen
\IEEEcompsocitemizethanks{\IEEEcompsocthanksitem CerCo, CNRS UMR 5549, Universit\'e de Toulouse and ANITI, Artificial and Natural Intelligence Toulouse Institute
\IEEEcompsocthanksitem Correspondence: rufin.vanrullen@cnrs.fr}
}

\date{}

\definecolor{blueColor}{RGB}{23, 179, 242}
\definecolor{greenColor}{RGB}{54, 181, 73}
\definecolor{redishColor}{RGB}{204, 51, 99}
\definecolor{orangeColor}{RGB}{252, 153, 13}
\definecolor{pinkColor}{RGB}{225, 164, 230}
\definecolor{purpleColor}{RGB}{129, 22, 186}

\newcommand{\ptr}{\textcolor{blueColor}{$P_\text{tr}$}}
\newcommand{\pcont}{\textcolor{purpleColor}{$P_\text{cont}$}}
\newcommand{\ptrdcy}{\textcolor{pinkColor}{$P_\text{tr}\ \allowbreak\&\ \allowbreak P_\text{dcy}$}}
\newcommand{\ptrcy}{\textcolor{greenColor}{$P_\text{tr}\ \allowbreak \&\ \allowbreak P_\text{cy}$}}
\newcommand{\ptrcont}{\textcolor{redishColor}{$P_\text{tr}\ \allowbreak \&\ \allowbreak P_\text{cont}$}}

\newcommand{\ptrcontdcycy}{\textcolor{orangeColor}{$P_\text{tr}\ \allowbreak \&\ \allowbreak P_\text{cont}\ \allowbreak \&\ \allowbreak P_\text{dcy}\ \allowbreak \&\ \allowbreak P_\text{cy}$}}

\colorlet{colorchangebg}{black!2}
\colorlet{colorchangeframe}{black!20}

\newenvironment{changes}{}{}

\newenvironment{blockChanges}{}{}

\begin{document}

\maketitle

\begin{abstract}
    Recent deep learning models can efficiently combine inputs from different modalities (e.g., images and text) and learn to align their latent representations, or to translate signals from one domain to another (as in image captioning, or text-to-image generation). However, current approaches mainly rely on brute-force supervised training over large multimodal datasets. In contrast, humans (and other animals) can learn useful multimodal representations from only sparse experience with matched cross-modal data. Here we evaluate the capabilities of a neural network architecture inspired by the cognitive notion of a ``Global Workspace'': a shared representation for two (or more) input modalities. Each modality is processed by a specialized system (pretrained on unimodal data, and subsequently frozen). The corresponding latent representations are then encoded to and decoded from a single shared workspace. Importantly, this architecture is amenable to self-supervised training via cycle-consistency: encoding-decoding sequences should approximate the identity function. For various pairings of vision-language modalities and across two datasets of varying complexity, we show that such an architecture can be trained to align and translate between two modalities with very little need for matched data (from 4 to 7 times less than a fully supervised approach). The global workspace representation can be used advantageously for downstream classification and cross-modal retrieval tasks and for robust transfer learning. Ablation studies reveal that both the shared workspace and the self-supervised cycle-consistency training are critical to the system's performance. 
\end{abstract}


\IEEEpeerreviewmaketitle


\input{sections/1_introduction}
\input{sections/2_related_work}
\input{sections/3_definitions}
\input{sections/4_modalities}

\input{sections/5_model}
\input{sections/6_experiments}

\input{sections/7_retrieval}
\input{sections/8_discussion}

\section*{Acknowledgments}
\noindent This research was funded by an ANITI Chair (ANR grant ANR-19-PI3A-004) and an ERC Advanced Grant GLoW (grant 101096017) to RV, an ANR grant COCOBOT (ANR-21-FAI2-0005), a Region Occitanie grant COCOPIL, and a ``D\'efi-cl\'e Robotique centr\'ee sur l'humain'' PhD-thesis grant to LM.
This work was performed using HPC resources from CALMIP (Grant 2020-p20032).
The authors would like to thank Alexandre Arnold for creating and sharing the WeBots environment used for the Factory dataset.
\bibliography{zotero_refs,refs}
\bibliographystyle{IEEEtran}

\clearpage
\appendices
\setcounter{page}{1}
\begin{center}
\textbf{\large Supplementary Materials}
\end{center}
\input{appendices/code}
\input{appendices/architectures}
\input{appendices/language_domain_generation}
\input{appendices/odd_one_out}

\end{document}

%% file: sections/1_introduction.tex
\section{Introduction}
\IEEEPARstart{H}{umans} learn about the world from various sources: images when looking around, language describing objects and their properties, sounds from the environment or from conversations, etc. These diverse inputs come together, sometimes asynchronously, to build a joint representation of the external world. Thanks to this multimodal convergence, human language is \emph{grounded} in the sensory environment and conversely, sensory perception is semantically \emph{grounded} by its relation to language~\cite{harnad1990symbol}.

Recent works have shown the importance of training deep learning models using several modalities such as vision and language. Paired inputs across modalities can be leveraged as natural and readily available annotations~\cite{radford_learning_2021}; they can be used as semantic constraints to train zero-shot learning models~\cite{frome2013devise, xian_zero-shot_2020} and more capable and general systems~\cite{lu_unified-io_2022}, or to infuse each modality with additional semantic knowledge (i.e., multimodal grounding)~\cite{silberer_grounded_2012, kiela_multi-_2015, pham_found_2019}. The performance of these models still heavily depends on the availability of large-scale paired multimodal datasets (400 million image-caption pairs used in CLIP~\cite{radford_learning_2021}, more than 4 billion pairs for CoCa~\cite{yu_coca_2022} or DALL·E 2 \cite{ramesh_hierarchical_2022}). This trend of brute-force supervised training on ever larger datasets has led to an impressive boost in the models' performance and to the emergence of new abilities in recent large AI models~\cite{wei_emergent_2022}. But it departs from the more frugal learning strategies adopted by the human brain. In addition, 
recent studies have shown that current multimodal networks sometimes fail to improve upon unimodal networks, i.e. they do not always achieve proper multimodal grounding \cite{devillers_does_2021,zhai_lit_2022}.

Here, we explore the capabilities of a multimodal system taking inspiration from the cognitive science theory of the Global Workspace (GW)~\cite{baars_cognitive_1993,dehaene_neuronal_1998}. GW Theory explains how different modalities in the human brain are integrated into a common shared representation, subsequently redistributed or \emph{broadcast} among the specialized unimodal modules (see section~\ref{section:GW}). We define essential properties that multimodal networks should verify (section~\ref{section:definitions}). In particular, translating signals between modalities, and aligning representations across modalities, are important and complementary abilities that should be jointly optimized. Yet most recent multimodal systems used either contrastive learning (e.g., CLIP \cite{radford_learning_2021}, TSM \cite{alayrac_self-supervised_2020}), which forces alignment of the modalities without preserving modality-specific information, or translation objectives (e.g. VirTex \cite{desai_virtex_2021}, ICMLM \cite{sariyildiz_learning_2020}), which do not provide a joint multimodal space for downstream tasks. Recently, the CoCa model \cite{yu_coca_2022} confirmed that using both translation and contrastive objectives for supervision could lead to significantly better performance. We show that our proposed GW-inspired architecture combines these two desired properties. Furthermore, to reduce annotations and encourage more frugal (and thus more human-like) learning, we also advocate for a semi-supervised learning setting, by adding unsupervised cycle-consistency objectives to the model. 
In short, we propose and evaluate a neural network architecture combining a global workspace with semi-supervision, in a bimodal setting where only few paired examples are available.

%% file: sections/2_related_work.tex
\section{Related Work}
As explained above, multimodal representation learning for neural networks is a vast and fast-growing research area, whose exhaustive coverage would require an extensive review well beyond the scope of the present paper. However, two specific features of our proposed model deserve a more in-depth treatment: unsupervised training via cycle-consistency, and the GW theory.

\subsection{Cycle-consistency}
The idea of using back-translations to synchronize two latent spaces has been introduced previously. In~\cite{kalal_forward-backward_2010}, the authors presented a forward-backward error to solve a visual point-tracking task. It consisted in predicting a forward trajectory of the tracked point in an image sequence and then predicting a reverse trajectory, considering the reversed image sequence. The two trajectories were then compared together.

More recently, a similar cycle-consistency principle was applied in NLP (Natural Language Processing) for unsupervised neural translation: language alignment is successful when the successive translation from language A to language B, then back-translation from B to A returns the original sentence~\cite{he2016, artetxe2018, Conneau2018, lample_unsupervised_2018}.
For instance,~\cite{lample_unsupervised_2018} used back-translations to optimize a sequence-to-sequence model with attention~\cite{bahdanau_neural_2015}, so that it could translate between two languages without ever having access to aligned multilingual corpora during training. Their training objective combined cycle-consistency with an adversarial loss to force the generation in each domain to match the actual language distributions.

Based on this logic, cycle-consistency was also used to synchronize multiple visual domains, i.e., unsupervised image-to-image translation~\cite{zhu_unpaired_2017, liu2017,yi2017}. For instance, in CycleGAN~\cite{zhu_unpaired_2017}, the authors trained two Generative Adversarial Networks (GANs) to generate images in the style of one specific domain, then used a cycle-consistency loss to synchronize the latent spaces of the GANs, so that an image from one domain (e.g. a horse) could be translated into the equivalent image in the other domain (e.g. a zebra).

From this point, it was not long until the technique was applied to multimodal use cases, such as text-to-image ~\cite{chaudhury2017,qiao2019,joseph2019} or touch-to-image translation~\cite{li2019}. For instance, \cite{pham_found_2019} applied cycle-consistency training to a multimodal (image, text, and sound) sentiment analysis task.
They showed that back-translations produced robust representations, and the model could deal with missing modalities during inference.
They used a hierarchical architecture, where two modalities are first aligned, then the third modality is aligned with the common latent space of the first two.
This architecture is asymmetric, and thus requires favoring some modalities over others. Upon trying different combinations, their best performance was obtained when first learning to translate between vision and text, then including audio.

Overall, it is clear that unsupervised learning via cycle-consistency is a powerful method to train multimodal systems. However, the technique is typically applied to multimodal translation tasks \emph{or} alignment tasks, but rarely to both; and it has never been combined with the GW architecture.

\subsection{Global Workspace}
\label{section:GW}

How the brain combines information from multiple modalities into a unified representation that can be flexibly re-used for a wide array of tasks is still the subject of active research. One prominent conjecture, however, is Baar's \textit{Global Workspace} (GW) theory \cite{baars_cognitive_1993,baars_global_2005}, extended into a neuronal framework by Dehaene et al.~\cite{dehaene_neuronal_1998}.
The theory comprises several components: a number of ``specialist'' modules, each independently processing one modality (visual stream, auditory stream, memory, motor, ...); an attention mechanism that determines the relevant specialist modules at each moment in time, based on both exogenous (saliency) and endogenous aspects (task, prior state); and a shared space with fixed capacity, the ``Global Workspace'' itself. Because of its fixed capacity, all modules cannot simultaneously access the workspace; this is why they must compete against each other through the attention mechanism. The winning modules transmit their information into the GW. Finally, the workspace representation is automatically broadcast to all modules. According to the theory, it is this broadcast of information that represents our inner experience, enabling multimodal grounding and flexible use for downstream tasks (including decision and action planning).
To illustrate his theory, Baars makes an analogy with a theater, where specialist modules are simultaneously the actors on a stage and the audience. While they are ``on stage'' (i.e., mobilized in the shared workspace) they can broadcast information to all other modules.


A recent opinion paper \cite{vanrullen_deep_2021} proposed that current working AI principles could already be used to implement this theory, and provided a step-by-step roadmap for this implementation.
Moreover, other implementations have recently been put forward. For instance, \cite{juliani_perceiver_2022} highlighted that the Perceiver architecture recently proposed by \cite{jaegle_perceiver_2021} could be used as a GW. Goyal et al.~\cite{goyal_coordination_2022} introduced an architecture for sharing information between modules (e.g. Transformer layers) that they explicitly labeled as a GW implementation.

The present work is not intended to compete with these prior systems, nor does it contend to offer a full implementation of Baar's theory. Instead, we make use of some prominent features of the GW theory (a unique, limited-capacity multimodal representation that can be broadcast and re-used for other tasks), while leaving others aside for future work. In particular, as we chose to work here with only two modalities, there is no need for us to consider attentional competition between modules (as only one domain can occupy the GW at each moment)--though this aspect will be important to examine in future studies.

\begin{changes}
    \subsection{Cross-modal Retrieval}
    Multimodal representations can be used for various applications. One of them is cross-modal retrieval, where the goal is to retrieve samples in one domain using a related query from another domain. The prevalent form is image--text retrieval, which consists in either retrieving the caption of an image, or the image that matches a specific description.
        
        \cite{hu_scalable_2019} uses a multimodal space that can be linearly projected (with a fixed matrix $P$) into the vectors of the classes pictured in the image. They use a combination of cycle consistency, and multi-class supervision. Alignment of modalities is achieved via multi-class supervision. This method is thus particularly well suited for a cross-modal model that ranks examples solely based on common classes.
        In~\cite{tian_multimodal_2022}, the authors use a combination of coupled DVAEs and a Fusion-Exchange VAE. They leverage disentanglement learning to extract modality-specific and modality-invariant information. They also introduce a Fusion-Exchange VAE to improve the alignment of modality-invariant features. Finally, they introduce the counter-intuitive cross-reconstruction strategy (CICR) where they learn to reconstruct the information of one modality with the decoder of the other modality. The model achieved SOTA results on image-text retrieval benchmarks. 
    
        We explore the performance of a global workspace-like architecture in a cross-modal retrieval task in section~\ref{sec:retrieval}. As opposed to the aforementioned methods, we focus on a setting with only a limited amount of paired image/text pairs. 
        
        The authors of~\cite{zhen_deep_2022} also trained their model with limited pairs of image/text. They tackle the issue that available paired samples could have different labels than additional unpaired samples. To address this, they use a discrimination loss with real labels whenever labels are available, and they use pseudo-labels (iteratively refined during training) as targets for unpaired data. They also use KL divergence to align feature representations of the image and text encoders, as done in other works \cite{yin_associate_2017, silva_playing_2020}.
        Our model differs by the use of cycle-consistency, and the contrastive loss (similar to CLIP) rather than a symmetric KL-divergence. 
        Another crucial difference is that we never use class-label information. Our models are trained only to learn a valid multimodal representation, i.e. one that supports vision-language alignment, translation and cycle-consistency objectives.
\end{changes}

%% file: sections/3_definitions.tex
\section{Problem statement}
\label{section:definitions}
\begin{figure}
    \begin{changes}
        \centering
        \includegraphics[width=0.90\linewidth]{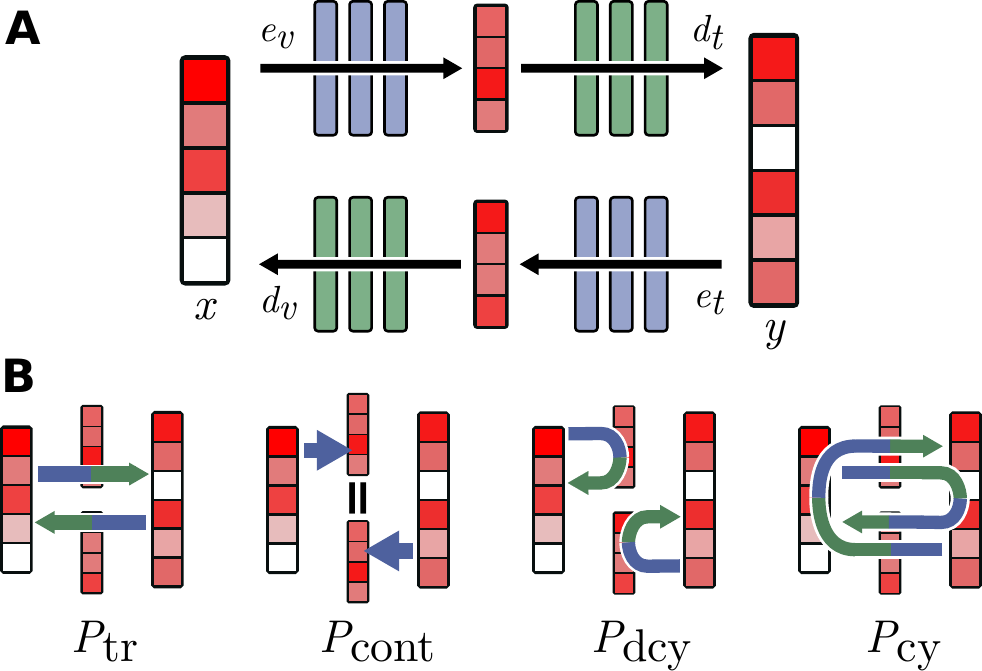}
        \caption{Panel A:\ generic bimodal network. Inputs can be from two modalities $x$ and $y$ (for instance, visual images and text captions). $e_v$ and $e_t$ are feed-forward neural networks that project each modality into a latent space; $d_v$ and $d_t$ are decoders that decode the latent space into the respective modality. (Note that this generic model aims at clarifying our definitions, it does not yet correspond to our GW architecture). Panel B:\ illustration of the primary and secondary desirable properties for multimodal systems. Each arrow shows a learned path to convert one latent vector into another. For instance in $P_{dcy}$ we can convert from one domain to itself via the central representation. Note that the four properties are not independent but can be causally related, as we describe in relations $R_1$ to $R_4$.}
        \label{fig:generic-multimodal}
    \end{changes}
\end{figure}
In this study, we will focus on a bimodal network receiving inputs from two modalities, such as vision (images) and text (captions). Figure~\ref{fig:generic-multimodal}A shows a diagram of a generic bimodal network, where we define visual and text encoders ($e_v$, $e_t$) and their respective decoders ($d_v$, $d_t$).

 $\{x_i\}$ represent a set of images, and $\{y_i\}$ their matched captions for $i\in\mathcal{S}$ a supervised set of matched examples. In the most general setting, our datasets could also include unmatched data samples, so we additionally define the supersets $\mathcal{U}_v$ and $\mathcal{U}_t$ for images and captions, respectively, such that $\mathcal{S} = \mathcal{U}_v \cap \mathcal{U}_t$. Images in $\mathcal{U}_v\backslash\mathcal{S}$ do not have a matched caption in the dataset, and similarly for captions in $\mathcal{U}_t\backslash\mathcal{S}$.

Multimodal networks can express different desirable properties. We define here two primary and two secondary properties that we believe represent fundamental behaviors that multimodal networks should possess.
The first primary property is translation, whereby it should be possible to translate each modality into the other one:
\begin{equation}
    \label{prop:tr}\tag{$P_\text{tr}$}
     \left\{
        \begin{aligned}
          & d_t(e_v(x_i)) = y_i \\
          & d_v(e_t(y_i)) = x_i
        \end{aligned}
    \right.\quad \forall i \in \mathcal{S}
\end{equation}
The second primary property is the contrastive encoding property, which forces the alignment of latent spaces for matched data:
\begin{equation}
    \label{prop:cont}\tag{$P_\text{cont}$}
     \left\{
        \begin{aligned}
          & e_v(x_i) = e_t(y_j) \quad \forall i=j \in \mathcal{S} \\
          & e_v(x_i) \neq e_t(y_j) \quad i\neq j
        \end{aligned}
    \right.
\end{equation}
As emphasized in the Introduction, these two properties correspond to two prominent objectives for multimodal learning in the literature; for example, text-to-image generation systems (like DALL·E 2~\cite{ramesh_hierarchical_2022} or Stable Diffusion~\cite{rombach_high-resolution_2022}) and image captioning models (like CoCa~\cite{yu_coca_2022}) are based on translation objectives, while CLIP~\cite{radford_learning_2021} relies on contrastive learning for vision-language representation alignment.

Although less often encountered in the literature, we believe that a bimodal network could also benefit from two additional ``secondary'' properties. We define the demi-cycle consistency, where $\forall (i,j) \in \mathcal{U}_v \times \mathcal{U}_t$:
\begin{equation}
    \label{prop:dcy}\tag{$P_\text{dcy}$}
     \left\{
        \begin{aligned}
          & d_v(e_v(x_i)) = x_i \\
          & d_t(e_t(y_j)) = y_j 
        \end{aligned}
    \right.
\end{equation}
and the (full) cycle consistency property, where $\forall (i,j) \in \mathcal{U}_v \times \mathcal{U}_t$:
\begin{equation}
    \label{prop:cy}\tag{$P_\text{cy}$}
     \left\{
        \begin{aligned}
          & d_v(e_t(d_t(e_v(x_i)))) = x_i \\
          & d_t(e_v(d_v(e_t(y_j)))) = y_j 
        \end{aligned}
    \right.
\end{equation}

These two properties can be seen as unsupervised versions of the primary properties, inspired by  the unsupervised language translation literature \cite{lample_unsupervised_2018, lample_phrase-based_2018}. Importantly, they are defined independently of the set $\mathcal{S}$ of paired samples, and remain valid even if $\mathcal{S}$ is an empty set (i.e., unsupervised learning).
\begin{changes}
    Figure~\ref{fig:generic-multimodal}B provides an illustration of these four properties.
\end{changes}

Note that our primary and secondary properties are not independent, and we can in fact highlight four relations between them that should be true for any matched samples in $\mathcal{S}$:
\begin{description}
    \item[$R_1$] \ref{prop:tr} $\Rightarrow$ \ref{prop:cy}
    \item[$R_2$] \ref{prop:tr} \& \ref{prop:cont} $\Rightarrow$ \ref{prop:dcy}
    \item[$R_3$] if $d_v$ or $d_t$ injective, \ref{prop:tr} \& \ref{prop:dcy} $\Rightarrow$ \ref{prop:cont}\footnote{$d_t(e_v(x_i)) \stackrel{P_\text{tr}}{=} y_i \stackrel{P_\text{dcy}}{=} d_t(e_t(y_i))$. Then using the injectivity of $d_t$: $e_v(x_i) = e_t(y_i)$. Similar proof if $d_v$ injective.}
    \item[$R_4$]\ref{prop:cont} \& \ref{prop:dcy} $\Rightarrow$ \ref{prop:tr}
\end{description}
The first two relations show that if we restrict ourselves to the set of matched samples, the secondary properties are automatically obtained if the primary properties are verified. Furthermore, the last two relations show that each primary property can follow from the other primary property combined with a secondary one (with some additional constraint on injectivity for $R_3$). Thus, we hypothesize that optimizing \emph{all four} properties in a single network could prove advantageous, since they will tend to reinforce each other as per relations $R_1$-$R_4$. Importantly, this means that the secondary properties could be used in an \emph{unsupervised} way to take advantage of unpaired data, and still enhance the network's primary properties.

How do the four properties relate to the Global Workspace architecture defined previously?
Within the scope of our study (i.e., for bimodal systems where attentional competition between modules is not required), a GW architecture must verify two criteria. First, it must have a common shared latent space across modalities, where a given input produces the same representation regardless of its modality of presentation; this corresponds exactly to property \ref{prop:cont}. Furthermore, the \emph{broadcast} aspect of GW theory implies that this shared space should be able to inform other modalities; thus, \ref{prop:cont} is necessary but not sufficient, as it only constrains the encoders $e_v$ and $e_t$ but not the corresponding decoders. Put another way, a model trained only for representation alignment, such as CLIP~\cite{radford_learning_2021}, cannot be considered to implement a GW. To train the decoders and permit broadcast, at least one of the other properties must also be optimized. In summary, a bimodal network can be said to include a GW if it verifies \ref{prop:cont} and at least one additional property in \{\ref{prop:tr}, \ref{prop:dcy}, \ref{prop:cy}\}.

Given these considerations, and our initial goal to study both the usefulness of a GW and the need for supervision in bimodal representation learning, we chose to focus our comparisons on four main models as shown in the first section of table~\ref{table:selected-models}. The models differ by the properties they are designed to optimize, such that two of them rely only on supervised training with paired bimodal samples, while the other two can also take advantage of additional unpaired data (semi-supervised training). Furthermore, two of them do not satisfy the criteria for a GW, while the other two do. Indeed, a model designed solely to optimize translation (\ref{prop:tr} and possibly its cycle-consistent version \ref{prop:cy}) can be thought of as operating with two entirely independent latent spaces, as illustrated in the middle of Figure~\ref{fig:generic-multimodal}A. It is only when imposing the representation alignment property \ref{prop:cont} (either directly, or indirectly via relation $R_3$) that the two latent spaces can be considered to work jointly as a unique shared space--the GW (see illustration in the middle of Figure~\ref{fig:gw:model_attr}). 

With this model selection (first section of Table~\ref{table:selected-models}), we can systematically investigate our two factors of interest, GW and semi-supervision. To train the models, we use the corresponding properties listed in the Table as our optimization targets. For evaluation, we measure the two primary desired properties (translation loss and contrastive loss) on a separate test set. For completeness, we also evaluate the secondary properties for each model after training. Finally, we check how the models perform on some downstream tasks.

\begin{table}
    \begin{blockChanges}
        \centering
        \resizebox{\linewidth}{!}{
            \begin{tabular}{@{}lcc@{}}
                \toprule
                \multirow{2}{*}{Model Properties} & \multirow{2}{*}{Semi-supervision} & ``Global Workspace''    \\
                                                  &                                   & (Alignment + Broadcast) \\
                \midrule
                \ptr{}                            & $-$                               & $-$                     \\
                \ptrcont{}                        & $-$                               & ++                      \\
                \ptrcy{}                          & ++                                & $-$                     \\
                \ptrcontdcycy{}                   & +++                               & +++                     \\
                \midrule
                \pcont{}                          & $-$                               & $\pm$~(no broadcast)    \\
                \ptrdcy{}                         & +                                 & +                       \\
                \bottomrule
            \end{tabular}
         }
        \caption{\label{table:selected-models} The compared models and their properties. All models share the same architecture, and differ only in terms of the primary and/or secondary properties that they are designed to optimize. The second column indicates whether each model relies on semi-supervised training; the third, whether the properties enforce the emergence of a Global Workspace, i.e. a combination of an aligned multi-modal representation space and the ability to broadcast the multimodal representation back to each modality. We mainly focus on the first four models in our experiments. Additionally, \pcont{} offers alignment, but not broadcast. \ptrdcy{} relies partly on semi-supervised learning, and could meet some requirements for a global workspace (according to relation~$R_3$). \ptrcy{} takes better advantage of semi-supervision, as the full cycle loss optimizes more parameters than demi-cycles. By combining all the losses, \ptrcontdcycy{} can simultaneously enforce all of the required properties, and is thus our proposed ``target'' model.}
    \end{blockChanges}
\end{table}

%% file: sections/4_modalities.tex
\section{Datasets}

\subsection{Simple Shapes}
To start with, we designed a multimodal dataset called ``Simple Shapes'' . The \textit{Simple Shapes} dataset is reminiscent of the 2D shapes dataset of \cite{higgins_beta-vae_2017}, but is extended with more varying attributes.
This dataset fits several objectives: firstly, we want an automated generation procedure, to obtain as many samples as needed.
It also allows us to control the number of annotations (i.e., matched samples) in the dataset.
Secondly, we want the modalities to overlap by representing the same content, so that we can train translation and alignment models between the modalities.
Thirdly, we want the models' architecture (and correspondingly, the data distribution) to be relatively simple so as to iterate quickly over several model training regimes for our analysis.

We consider two modalities: vision and language.

\subsubsection{Visual modality}
\begin{figure}
\centering
\includegraphics[width=\linewidth]{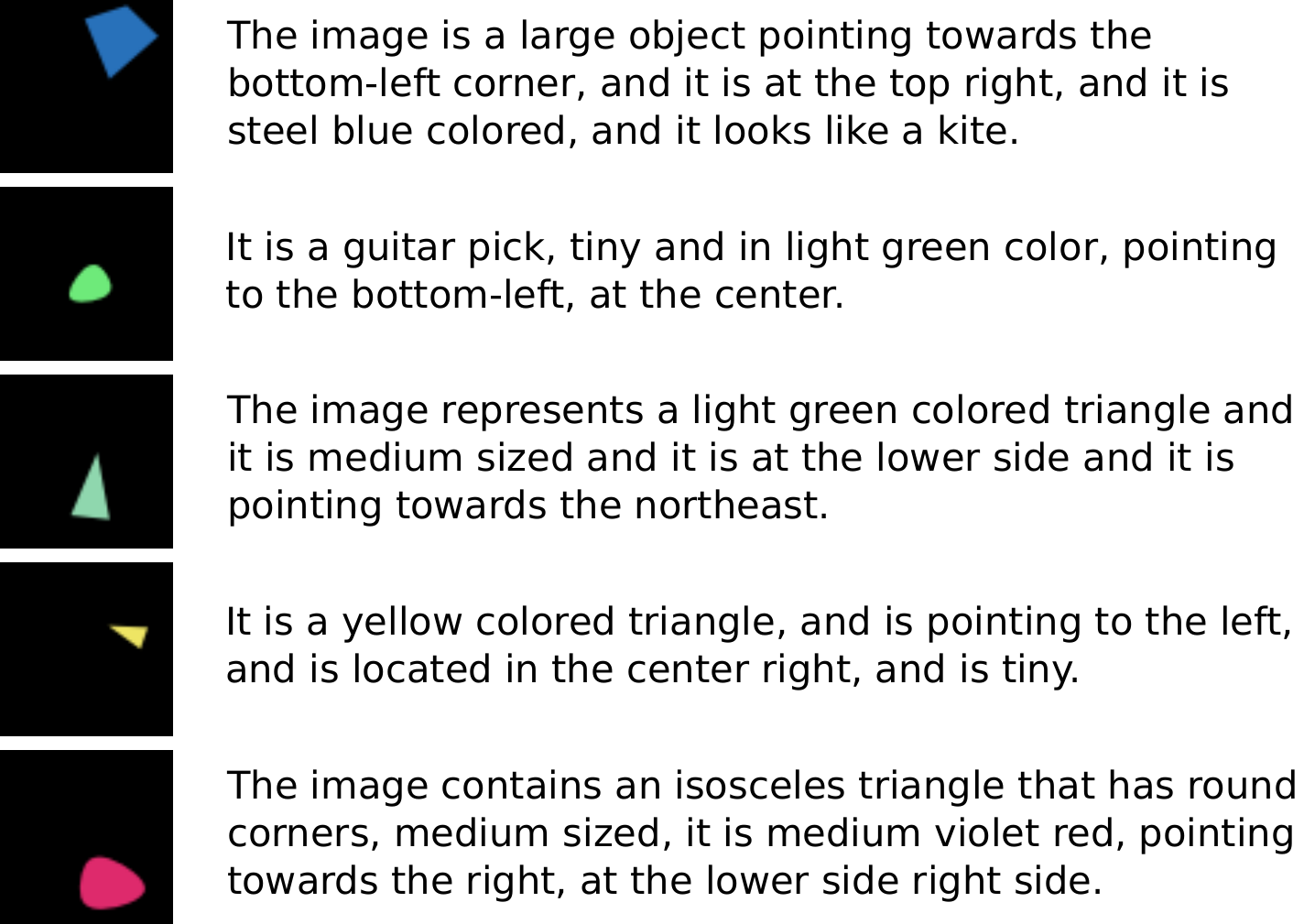}
\caption{Examples from the \textit{Simple Shapes} dataset. Each image contains a unique object of differing shape, color, rotation, size, and position. The image is paired with a natural language sentence describing the attributes.}
\label{fig:gw:dataset:visual_domain}

\end{figure}
For the visual domain, we create small images of $32\times 32$ pixels with a black background and a unique object fully visible in the frame.
Figure \ref{fig:gw:dataset:visual_domain} shows examples of the visual domain.
The object can be of 3 categories: an egg-like shape, an isosceles triangle, and a diamond. These categories were chosen so that the shape's orientation can be defined unambiguously.
Each object has several attributes which are sampled uniformly: a size $s\in[s_\text{min}, s_\text{max}]$, a location $(x, y)\in[\frac{s_\text{max}}{2}, 32-\frac{s_\text{max}}{2}[^2$ (we add a margin of size $s_\text{max}/2$ so that images of all sizes are completely in frame), a rotation $r \in [0, 2\pi[$, and an HSL (hue, saturation, lightness) color $(c_h, c_s, c_l)\in[0, 1]^2\times[l_\text{min}, 1]$, then translated into RGB (this ensures that images can always be seen on a black background, by setting a minimum lightness value).

\subsubsection{Language modality}
\paragraph{Proto-language}
First, we use a form of ``proto-language'', defined by the attributes and categories that were used to produce the images. It contains a 3-dimensional one-hot annotation for the class, two numerical values for the position ($x,y$), one for the size, 3 for the colors (in RGB), and we transform the angle value of the rotation into 2 values for its sine and cosine. We found that describing the angle in the cos/sin space yields better results, as it avoids the wrap-around discontinuity around $0$ or $2\pi$, which can lead to a significant error signal when using the mean-squared-error (MSE) loss. Besides, all the attributes are normalized to have a value between -1 and 1. This \emph{proto-language} modality is useful because it guarantees an exact and unique match between the descriptions of any data sample in the two modalities.

\paragraph{Natural language}
In addition, we implement a natural language modality to describe the visual aspects of the image in plain English. The text is automatically generated from the semantic (proto-language) vectors using a heuristic method described in appendix~\ref{appendix:generating-text}. Training with natural language adds complexity, as the attributes are quantized into words, meaning we go from a continuous distribution to a categorical one.
Moreover, natural language contains uncertainty since a single word can be used to capture a distribution instead of a specific range of attribute values: for instance, ``small'' and ``medium'' may be used to refer to the same object size depending on the person or context.
Using natural language also affords more liberty in the description of the shapes, the structure of the sentence, or even the vocabulary. This makes multimodal translation and alignment inherently more difficult, because different sentences can be used to describe the same object, and slightly different objects could be described by the exact same sentence. In other words, when using natural language, multimodal alignment is typically not bijective.

\begin{figure}
    \centering
    \includegraphics[width=\linewidth]
    {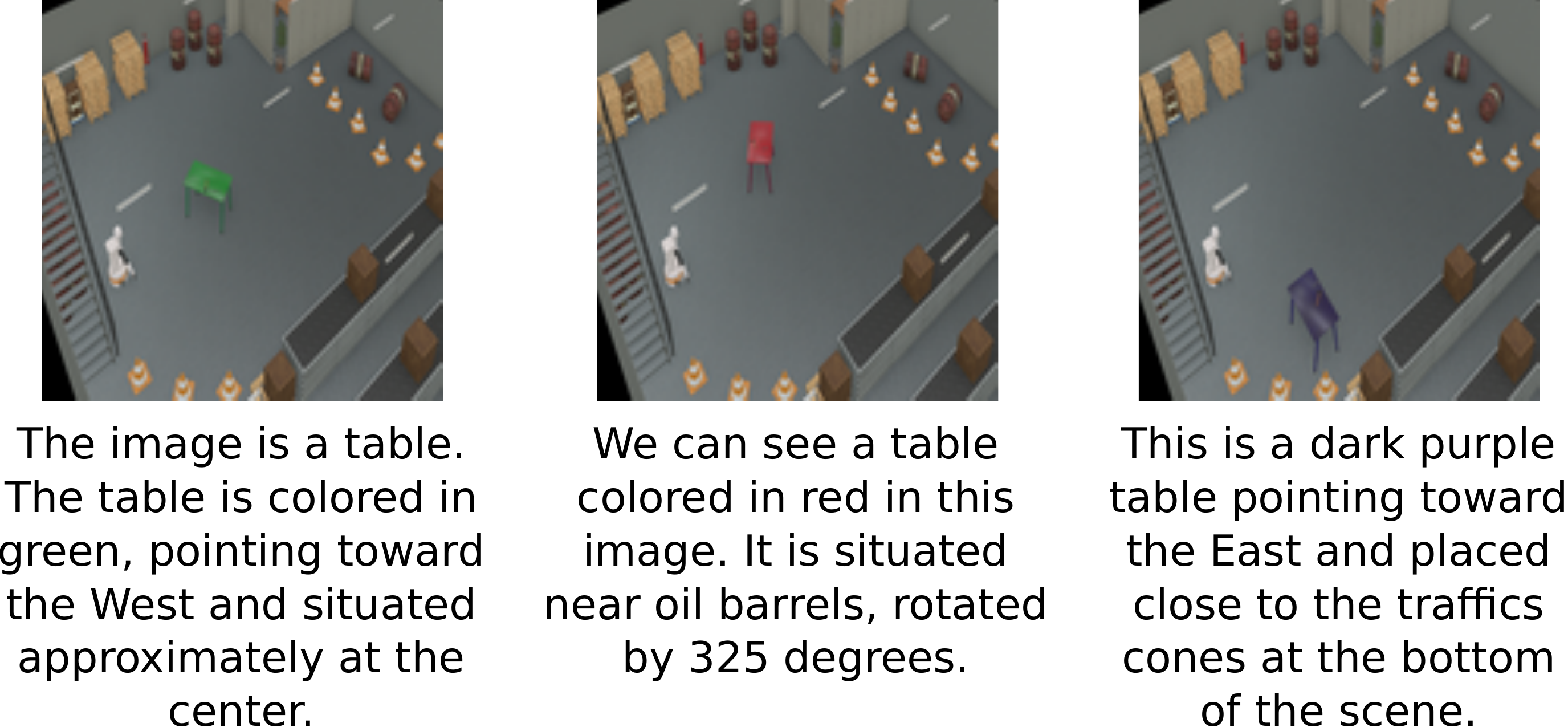}
    \caption{Examples from the \textit{Factory} dataset. Each image is taken from a fixed point of view; a table is randomly positioned in the environment, while the other objects (robot, cones, crates, conveyer belt...) remain in a fixed position. Each image can be associated with a ``proto-language'' description of the table's attributes (position, orientation, color), or with a natural language English description (as shown on the right).}
    \label{fig:gw:factory:visual_domain}
\end{figure}
\begin{changes}
    \subsection{Factory}
    
    We create a second synthetic dataset called \textit{Factory}, composed of $K=200,000$ image-text pairs. This dataset remains very close to the \textit{Simple Shapes} dataset in principle, but using more realistic 128 $\times$ 128 pixels images from a simulated robotic environment (defined using the Webots simulator \cite{Webots}). The scene viewpoint, overall layout and the position of most objects (robot, barrels, crates, cones, conveyer belt...) are fixed across images; only a table varies, with randomly chosen attributes: position $(x,y)$, orientation, color hue.
    Figure \ref{fig:gw:factory:visual_domain} (left) shows examples of images from the \textit{Factory} dataset.
    
    As in Simple Shapes, we can describe images using a ``proto-language'' (attribute vectors) or using natural language (English). The attributes describing the image in the proto-language contain two values $(x,y)$ for the position of the table (normalized between -1 and 1). For the table orientation, the angle $\theta$ around the z axis is transformed as $(\cos(2\theta), \sin(2\theta))$ (angle multiplied by 2 because of the table's symmetry modulo $\pi$). The table's color only varies in the (circular) Hue domain, so it is transformed as an angle: $(\cos(2\pi H), \sin(2\pi H))$.
    As for the \textit{Simple Shapes} dataset, we also generated natural language sentences describing each image, using a heuristic method based on the attributes (see Appendix). 
    Examples of generated sentences are shown in Figure \ref{fig:gw:factory:visual_domain} (right).

    \subsection{COCO Captions}

    Finally, to replicate our findings on a dataset with naturalistic images and human-written captions, we will evaluate on COCO Captions~\cite{chen_microsoft_2015}.
    The dataset contains 82,783 training, 10,000 validation and 30,504 test visual scenes with text descriptions. For each image, there are 5 captions describing the visual scene in natural language. We use the popular Karpathy split~\cite{karpathy_deep_2015} with 5,000 images for both validation and testing and use the remaining 30,504 images for additional training examples (``restval''). 
\end{changes}

%% file: sections/5_model.tex
\section{Model}
All of our models are connected to two modalities: vision and language (proto- or natural language). 
The corresponding specialist modules are pre-trained independently on their modality, and subsequently frozen when training the multimodal networks. 

\begin{figure}
\centering
\includegraphics[width=\linewidth]{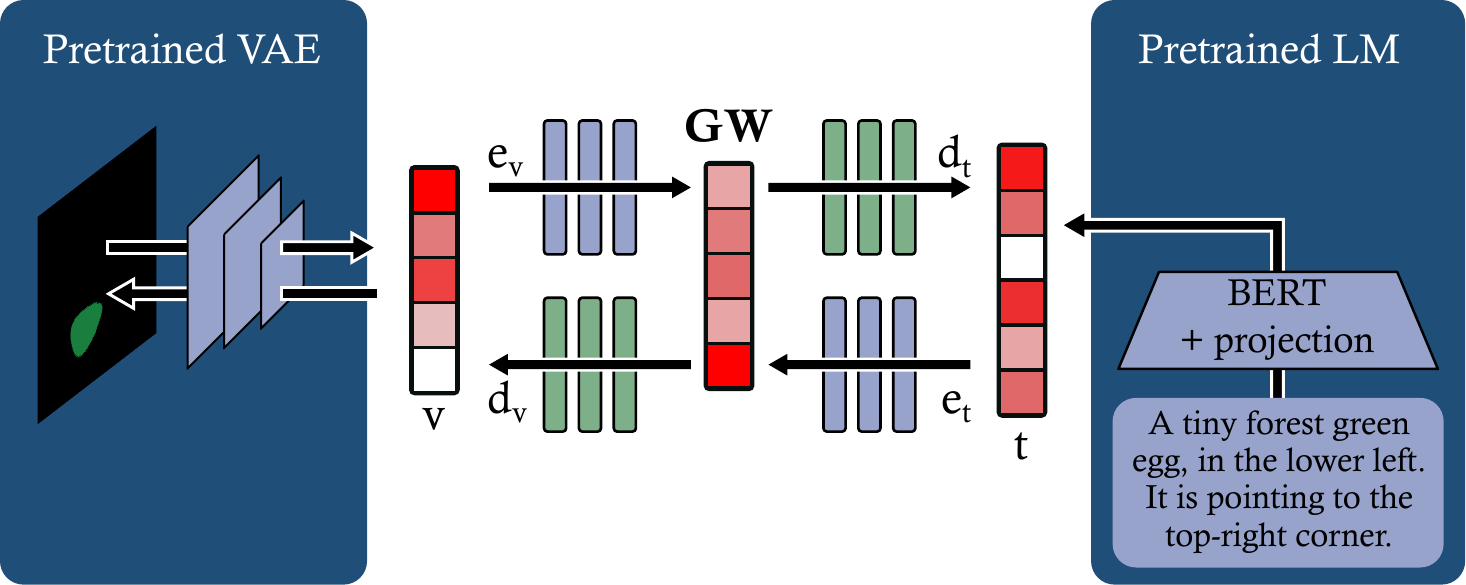}
\caption{Diagram of our Global Workspace architecture. Specialist modules for vision and language have a blue background. We use each modality's encoder ($e_v$ or $e_t$) to project data samples into a common latent space (GW), and the corresponding decoders ($d_t$ or $d_v$) to translate GW activations into the input domains. In this figure, we make the assumption that the model verifies property \ref{prop:cont}, thus having the GW representation shared across modalities.}
\label{fig:gw:model_attr}
\end{figure}

\subsection{Specialist modules}
\subsubsection{Visual domain}
For the visual specialist module, we use a $\beta$-VAE. The details of the architecture are provided in Appendix~\ref{appendix:architecture}.
We chose a VAE \cite{kingma_auto-encoding_2014} instead of a regular auto-encoder, to take advantage of the $\mathcal{N}(0,1)$ normally distributed latent space, which may be helpful for multimodal alignment and translation.
Moreover, we use a $\beta$-VAE \cite{higgins_beta-vae_2017} as opposed to a regular VAE to obtain a more regularized latent space.
Indeed, the latent space of a $\beta$-VAE has more disentangled dimensions, which again could assist the translation and alignment between the modalities.
We train the VAE on the visual domain data only, using a 12-dimensional latent space, and use $\beta=0.1$, a value chosen to optimize the reconstruction quality while keeping a normal latent vector to allow sampling.

\begin{changes}
    For the Factory dataset, given that most of the background is identical across images, and only the table varies, a standard VAE architecture (with a pixel-based reconstruction loss) was found to be inappropriate, as it tended to favor the background at the expense of the table details. To encourage our vision module to properly encode the table, we used a learnable tensor image $X_0$ that was subtracted from the image before entering the VAE, and added back to the VAE reconstruction. This way, the learnable tensor would capture fixed elements of the scene, while the VAE could focus on changing elements. Details of this architecture and examples of encoded and reconstructed images are given in Appendix~\ref{appendix:architecture}.
\end{changes}

\subsubsection{Proto-language domain}
For the proto-language, we directly use a vector containing the concatenated attributes, normalized between -1 and 1. 

\subsubsection{Language domain}
For the text domain, we use a pretrained BERT model to encode the natural language sentences into a latent vector. This high-dimensional vector (768-D) encodes extra information (e.g. syntactic or grammatical) that cannot be aligned with or translated into the visual domain. Thus, to simplify the task and maintain reasonably compact latent representations, we use another VAE (see architecture details in Appendix~\ref{appendix:architecture}) to project the BERT vector into a smaller (12-dimensional) latent representation; in addition to the standard VAE training objectives (reconstruction of the initial BERT vector, KL loss), we add an attribute- and grammar-prediction head (see architecture details in Appendix~\ref{appendix:architecture}) and use its prediction loss in the optimization.

\begin{changes}
    For Factory, the latent space dimensionality was increased from 12 to 20, and we added layers for the regression of attributes (see Appendix~\ref{appendix:architecture}).
\end{changes}

\subsection{Objectives of the multimodal system}
We train a multimodal system structured around a GW, i.e., an intermediary space that allows unimodal latent spaces to communicate (see Figure~\ref{fig:gw:model_attr}).
To connect the modalities to the GW, we use one encoder $e_m$ for each connected modality $m$. Each encoder projects the unimodal latent space into the workspace. Moreover, a decoder $d_m$ translates the GW representations back to the domain's unimodal latent space.
In our experiments, $e_m$ and $d_m$ are 4-layers feedforward models with a 12-dimensional input, 256-dimensional hidden layers, and a 12-dimensional output.

\begin{changes}
    In Factory, we used 4 hidden layers of size 512 each for the encoders and decoders. The GW latent space itself had 10 dimensions (see Appendix~\ref{appendix:architecture}).
\end{changes}

As explained in Section~\ref{section:definitions} and Table~\ref{table:selected-models}, our full GW system is intended to both align visual and language inputs into a common latent space, and to translate inputs from one domain into the other. These different objectives correspond to our ``primary properties'' \ref{prop:cont} and \ref{prop:tr}, and are enforced in our system via distinct training losses. In addition, our training setting is semi-supervised, i.e., we employ both supervised and unsupervised losses. The unsupervised objectives (corresponding to our ``secondary properties'' \ref{prop:dcy} and \ref{prop:cy}) are based on the cycle-consistency principle, and can be thought of as regularization terms that can accelerate the optimization process and improve the model's generalization.
We now describe the different loss components in detail.

\subsubsection{Translation loss}
The first primary property \ref{prop:tr} is handled by the translation loss, where we predict one domain from the other (and vice-versa), using pairs of matching examples $(x_i, y_i)$ with $i\in\mathcal{S}$.
First, let us define the translation function from a domain to another as:
\begin{equation}
\label{eq:gw:translation_fn}
\tau_{v \rightarrow t}(x_i) = d_t\left( e_v(x_i)\right), \quad \forall i \in \mathcal{S}
\end{equation}
Now we can express the translation objective:
\begin{equation}
\label{eq:gw:translation_objective_one}
\mathcal{L}_{\text{tr}}^{v\rightarrow t} = \frac{1}{|\mathcal{S}|}\sum_{i\in\mathcal{S}} \ell_{MSE}\left(\tau_{v \rightarrow t}(x_i), y_i\right)
\end{equation}
where $\ell_{MSE}$ measures the Euclidian distance in the target space\footnote{In the case of the proto-language, we combine an MSE loss for the size, rotation, location and color, with a cross-entropy loss for the prediction of the object category.}.
The full translation loss is the average of the losses from both  ``directions'':
\begin{equation}
\label{eq:gw:translation_objective}
\mathcal{L}_{\text{tr}} = 0.5(\mathcal{L}_{\text{tr}}^{v\rightarrow t} + \mathcal{L}_{\text{tr}}^{t\rightarrow v})
\end{equation}

\subsubsection{Contrastive loss}
\ref{prop:cont} is optimized using the contrastive loss:
\begin{equation}
\label{eq:gw:contrastive_objective}
\mathcal{L}_\text{cont} = -\sum_{i\in \mathcal{S}}\sum_{j\in \mathcal{S}} \ell_{CE}\left(\frac{e_v(x_i) \cdot e_t(y_j)}{\|e_v(x_i)\| \|e_t(y_j)\|}, \mathds{1}_{i=j}\right)
\end{equation}
where $\mathds{1}_{i=j} = 1$ when $i=j$ and $0$ otherwise, and
$$\ell_{CE}(p, q) = q\log(p) + (1-q)\log(1-p)$$

In a nutshell, the contrastive loss ensures that the normalized dot product between matching exemplars across modalities is close to $1$ (i.e., that the latent vectors are aligned), but close to $0$ for non-matching exemplars (i.e., the latent vectors are orthogonal). The translation objective (eq. \ref{eq:gw:translation_objective}) optimizes both the encoders and decoders, which allows us to use it as a standalone objective in some of our ablated models. 
On the other hand, contrastive learning (eq. \ref{eq:gw:contrastive_objective}) only optimizes the encoders. 
When learning with the contrastive loss, we thus require at least one additional objective (see Table~\ref{table:selected-models}, right).

\subsubsection{Cycle-consistency}
Unsupervised objectives (corresponding to the secondary properties of section~\ref{section:definitions}) are split into (full) cycle~\ref{prop:cy} and demi-cycle~\ref{prop:dcy} consistency losses. They are defined over the entire training set ($\mathcal{U}_{v}$ or $\mathcal{U}_{t}$) rather than only the paired data $\mathcal{S}$.

To introduce the full cycle-consistency loss (~\ref{prop:cy}), we first define the cycle function by combining two translations:
\begin{equation}
\label{eq:gw:cycle_fn}
c_{v}(x_i) = \tau_{t \rightarrow v}\left(\tau_{v \rightarrow t}\left(x_i\right)\right), \quad \forall i \in \mathcal{U}_{v}
\end{equation}
Now let us define the loss over (for instance) the visual modality as:
\begin{equation}
\label{eq:gw:full_cycle_objective}
\mathcal{L}_{\text{cy}}^{v}= \frac{1}{|\mathcal{U}_{v}|}\sum_{i\in\mathcal{U}_{v}}\ell_{MSE}\left(c_v(x_i), x_i\right)
\end{equation}
So that the full objective when training with two modalities is:
\begin{equation}
\mathcal{L}_{\text{cy}} = 0.5(\mathcal{L}_{\text{cy}}^{v} + \mathcal{L}_{\text{cy}}^{t})
\end{equation}

\subsubsection{Demi-cycle-consistency}
Finally, the demi-cycle consistency loss over (for instance) the visual domain is defined by:
\begin{equation}
\label{eq:gw:demi_cycle_objective_domain}
\mathcal{L}_{\text{dcy}}^v = \frac{1}{|\mathcal{U}_{v}|}\sum_{i\in \mathcal{U}_{v}} \ell_{MSE}\left(\tau_{v \rightarrow v}(x_i), x_i\right)
\end{equation}
Note that evaluating $\tau_{v \rightarrow v}$ defined in eq.~\ref{eq:gw:translation_fn} with the same source and target domains amounts to performing a demi-cycle $d_v(e_v(x))$. 
Then, we obtain the full loss~\ref{prop:dcy} by averaging the two possible demi-cycles:
\begin{equation}
\label{eq:gw:demi_cycle_objective}
\mathcal{L}_{\text{dcy}} = 0.5(\mathcal{L}_{\text{dcy}}^{v} + \mathcal{L}_{\text{dcy}}^{t})
\end{equation}

Both unsupervised objectives serve a different purpose: the cycle-consistency loss ensures that the two domains are synchronized by translation (i.e. $\tau_{t\rightarrow v}$ and $\tau_{v\rightarrow t}$ are mutually inverse functions); the demi-cycle consistency ensures that $d_v$ (resp. $d_t$) and $e_v$ (resp. $e_t$) are inverse functions of one another and thus forces the global workspace to coordinate the representations of the domains.

\subsection{Is supervision necessary?}
Cycle-consistency and demi-cycle consistency losses are intended to align the encoders and decoders (and the resulting translations) across the two modalities, so that each representation can be consistently inverted. However, aligning two domains without any additional constraint is an intrinsically ambiguous problem. Indeed, if there exists at least one bijection from one domain to the other, then by randomly permuting the samples we can produce many other equally valid bijections.
For example, imagine mapping a letter domain $\{a, b, c\}$ onto a number domain $\{1, 2, 3\}$. Any one-to-one mapping is technically correct, but we might want to enforce that $a \leftrightarrow 1, b \leftrightarrow 2$, and $c \leftrightarrow 3$. Unsupervised learning techniques cannot directly enforce this constraint, but a small amount of supervision (i.e., labelled examples) might be sufficient. From the example above, if we additionally provide that $a$ maps to $1$ and $c$ maps to $3$, then the ambiguity is completely removed. This explains why supervision can be important for our multimodal learning problem, i.e. why the translation and/or contrastive losses are needed. However, just like in our $\{a, b, c\}$ example, the number of labelled samples that are needed may be relatively small. The need for supervision will be explicitly quantified in our experiments.

\subsection{Combining the objectives}
The final loss function is a combination of the four different objectives with different weights:

\begin{equation}
\label{eq:gw:final_objective}
\mathcal{L} = \alpha_{\text{tr}} \mathcal{L}_{\text{tr}} + \alpha_{\text{cont}} \mathcal{L}_{\text{cont}} + \alpha_{\text{cy}} \mathcal{L}_{\text{cy}} + \alpha_{\text{dcy}} \mathcal{L}_{\text{dcy}}
\end{equation}

In our implementation, the shared workspace is implicit, i.e. it emerges from the chosen training objectives. For example, training a model with only the translation loss or the cycle loss does not truly produce a GW. Indeed, they do not force the output of the encoders to project into a similar space, effectively resulting in a situation akin to the illustration in Figure~\ref{fig:generic-multimodal}. 
On the contrary, the contrastive loss (and to some extent, the demi-cycle loss, as per relation $R_3$) explicitly forces the encoders' output to be aligned across modalities, effectively resulting in the situation illustrated in Figure~\ref{fig:gw:model_attr}. Thus, by setting the weight of some of the loss terms to zero in equation~\ref{eq:gw:final_objective}, we can easily modulate the effective architecture of the model, and probe the functional relevance of our two factors of interest, GW and semi-supervision, as proposed in Table~\ref{table:selected-models}.

%% file: sections/6_experiments.tex
\section{Experiments}
\label{section:experiments}

\begin{figure*}
    \centering
    \includegraphics[width=\linewidth]{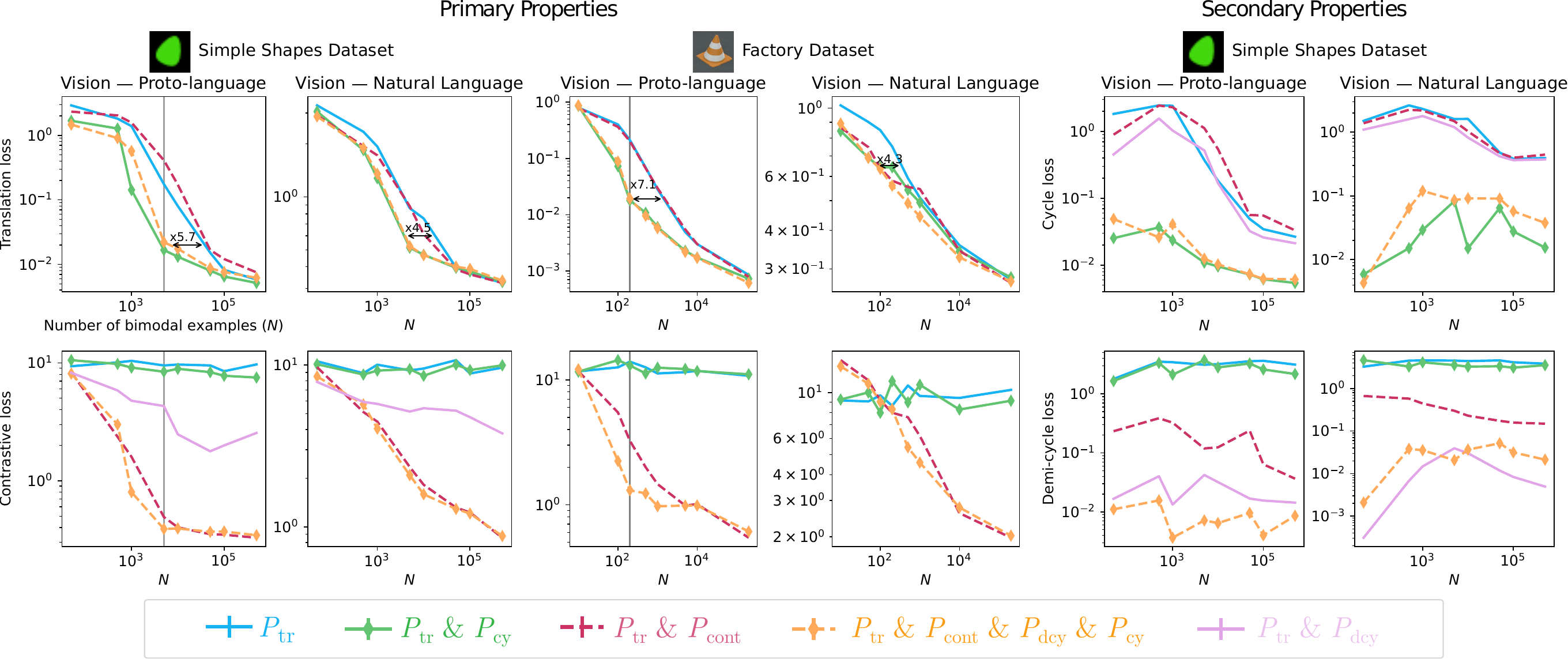}
    \caption{The left panel evaluates the primary properties (translation and contrastive alignment); the right panel assesses the secondary properties (cycle and demi-cycle consistency). Each point in each graph is a different model trained until convergence, using a particular number of matched bimodal examples $N$ (x-axis). Dashed lines correspond to GW models, and curves with markers are semi-supervised models. \ptrdcy{} was included as a way to assess relation $R_3$. The first and second rows on the left display the test translation and contrastive losses of the selected models, respectively. The first and second rows on the right show the test cycle and demi-cycle losses, respectively. Columns refer to different language modalities (proto-language or natural language). (The vertical gray line in the leftmost column marks the chosen value of $N$ that will be used later to assess the influence of the total number of unsupervised data samples.)}
    \label{fig:properties_results}
\end{figure*}

Our main experiments involve the \textit{Simple Shapes} dataset, composed of $K=500~000$ image-text pairs. (In the next section, we will also validate our findings on another dataset.)
We report all translation/alignment experiments with both language types, i.e. vision $\leftrightarrow$ proto-language training, and vision $\leftrightarrow$ natural language training. To characterize the need for labelled data across the two domains, we artificially split our training dataset into aligned data $\mathcal{S}$ (both domains are available) and unaligned data $\mathcal{U}_v$ and $\mathcal{U}_t$ (only one of the two domains is available), by randomly selecting $N \le K$ matched examples from our original set. 
\begin{changes}
    During training, we use the pool of $N$ matched samples and another of $2K$ unmatched samples (for each modality, $K - N$ that are truly unmatched, plus $N$ of $\mathcal{S}$ that have been artificially decoupled).
\end{changes}
We create batches by drawing samples from these two pools with equal probability, thus having equal numbers of paired and unpaired data at every learning step, regardless of the value of $N$. (But of course, for small $N$, the same paired data will re-occur more often during training, while the training diversity will increase with increasing $N$).

We optimize the model with different values for $N\in\{50,\allowbreak 100,\allowbreak 500,\allowbreak 1000,\allowbreak 5000,\allowbreak 10~000,\allowbreak 50~000,\allowbreak 100~000,\allowbreak 500~000\}$ using the loss in equation \ref{eq:gw:final_objective}. For the loss coefficients, we fix $\alpha_\text{tr}=1$. For $\alpha_\text{cont}$, $\alpha_\text{cy}$, and $\alpha_\text{dcy}$ we either do not use the loss (coefficient set to 0, reflecting the model choice in Table~\ref{table:selected-models}), or choose among three possible values \{0.1, 1, 10\} the one yielding the best result for both translation and contrastive objectives\footnote{We choose the coefficient that minimizes a weighted average of the translation and contrastive losses. To select the weights of the weighted average, we train one model with only a translation loss and one with only a contrastive loss, and we select the weights so as to equalize the translation/contrastive losses at the end of training.}.
\begin{changes}
    Retrospectively, we found that keeping a coefficient of 1 for translation, cycle and demi-cycle, and a low coefficient for the contrastive loss works in most of our experiments. Indeed, we used these hyperparameters for the COCO experiments instead of the grid search strategy (using a contrastive coefficient of 0.05).
\end{changes}
Regardless of the training regime (i.e., the value of $N$), we evaluate all of our models on the same independent test set of $1000$ images and matching descriptions.

\subsection{Primary properties}

\subsubsection{SimpleShapes}
We start by evaluating the models on the two primary properties \ref{prop:tr} and \ref{prop:cont}.
We report the value of the translation and contrastive test losses as a function of $N$ in figure~\ref{fig:properties_results} (left). Each point in the graph is a model trained until convergence. To facilitate comparisons, the same matched examples are used for all models with the same value of $N$. 

To relate the curves with the models listed in table~\ref{table:selected-models}, we use the following conventions: curves with solid lines (resp. with dashed lines) correspond to models with  the Global Workspace property (resp. without the property); curves with no marker (resp. with diamond markers) correspond to models without the semi-supervision property (resp. with supervision); the color of the curves also matches the text color in table~\ref{table:selected-models}. 

\begin{changes}
    Note that we consistently test all 4 baseline models for both translation and contrastive objectives, even though some models are not explicitly trained to optimize them. This allows to verify our four proposed relations $R_1$ to $R_4$ (e.g.\ \ptrdcy{} is not trained with a $P_\text{cont}$ objective, but we can still observe alignment, in accordance with relation $R_3$).
\end{changes}

As expected, we observe in figure~\ref{fig:properties_results} that increasing the amount of aligned examples improves all of the models' translation performance. However, the different models improve at distinct rates. We observe that semi-supervised models (with diamond markers) require significantly fewer annotations to obtain the same results. For instance, a semi-supervised model trained with $N=5000$ matching pairs \{image, proto-language annotation\} performs approximately as well as a fully supervised model trained with $N=30,000$ matching pairs---an $\sim6$-fold improvement. Similarly, a model trained to translate between images and natural language captions using semi-supervision requires 4.5 times fewer matching examples than the equivalent fully supervised model. In short, the advantage of semi-supervision is readily apparent for the translation property.

Looking at the contrastive property, we see that only the models equipped with a GW perform well (dashed lines). 
Indeed, without a GW, the encoded representations from each domain do not need to be aligned with each other (see illustration in figure~\ref{fig:generic-multimodal}).
For example, in a model trained only for translation (blue curve), encoding/decoding sequences (such as $d_t(e_v(x))$ or $d_v(e_t(y))$) can be viewed as direct (merged) translation functions (that is, $\tau_{v \rightarrow t}(x)$ or $\tau_{t \rightarrow v}(y)$, see eq~\ref{eq:gw:translation_fn}), so there is no intermediate latent multimodal representation to speak of.

Measuring the contrastive loss also allows us to evaluate the validity of relation~$R_3$: optimizing translation and demi-cycle losses (without explicitly optimizing the contrastive loss) should be sufficient to achieve property \ref{prop:cont}. Theoretically, this relation holds for samples in $\mathcal{S}$, and given that the encoders are injective functions. To assess whether this relation can generalize to the test set, we trained additional models with only translation and demi-cycle losses, and plot the resulting curve (in pink) on figure~\ref{fig:properties_results}. We see that combining translation and demi-cycle losses can help align the visual and text multimodal representations, compared to a translation-only model. However, the resulting alignment is partial, i.e. weaker than when explicitly optimizing the contrastive loss. This likely indicates that the trained models cannot perfectly generalize to the test data, and/or that our encoders are not stricly injective functions.

In conclusion, semi-supervision was shown to be beneficial to learn the translation primary property~\ref{prop:tr}, by decreasing the need for annotated data. In addition, including a GW in the architecture permits multimodal alignment in the intermediate latent space, thereby satisfying property~\ref{prop:cont}. 
Overall, the best model to jointly satisfy our two primary properties is the one combining a GW architecture with self-supervised training (\ptrcontdcycy{}).

\begin{changes}
    \subsubsection{Factory}
    
    As done with the \textit{Simple Shapes} dataset, we used the loss in equation \ref{eq:gw:final_objective} to optimize the various models with different values for $N\in\{50,\allowbreak 100,\allowbreak 200,\allowbreak 300,\allowbreak 500,\allowbreak 5000,\allowbreak 10~000,\allowbreak 200~000\}$. We used a translation loss coefficient $\alpha_{tr} = 1$ for all experiments. The cycles, demi-cycles and contrastive coefficients were chosen among $\alpha_{cy} \in \{1, 2, 4, 8, 10\}$, $\alpha_{dcy} \in \{1, 2, 4, 8, 10\}$, $\alpha_{cont} \in \{0.0005, 0.005\}$ so as to jointly optimize the primary properties~\ref{prop:tr} and~\ref{prop:cont}.
    
    For these validation experiments, we focused on the two primary properties~\ref{prop:tr} and~\ref{prop:cont}. Figure \ref{fig:properties_results} shows the test performance of each model trained with a certain amount of bimodal matched examples ($N$). Each point corresponds to one model trained until convergence; for a given value of $N$, all compared models used the same training data. Training batches were generated in the same way as described in section~\ref{section:experiments} (but with $K=200,000$ instead of $K=500,000$).
    
    Overall, the results are similar to the ones obtained with the \textit{Simple Shapes} dataset. When the number of bimodal matched examples (N) increases, the  translation loss of all models decreases, for both proto-language and natural language domains. There is also a performance gap between models trained only with supervision (blue and red curves) and semi-supervised models trained also with unimodal (unmatched) data (orange and green curves). For the proto-language translation the greatest effect of semi-supervision is visible at $N=200$. At this point, the semi-supervised models have the same translation loss as the supervised models trained with $\sim7$ times more bimodal paired examples. A qualitatively similar but more modest effect of semi-supervision is also observed for the natural language translation (with e.g. a gap of x4.3 between the model trained with translation loss only, and the one trained with an additional cycle loss).
    
    As previously, multimodal alignment (as measured by the contrastive loss in Figure \ref{fig:properties_results}, bottom) only emerges for models including a GW (red and orange curves, \ptrcont{} and \ptrcontdcycy{}). If no alignment constraint is applied on the encoders (with contrastive and/or demi-cycle losses), then the encoded visual and linguistic data are not aligned (blue and green curves, \ptr{} and \ptrcy{}). The benefits of semi-supervision can also be observed here for the contrastive loss. In the lower part of Figure \ref{fig:properties_results} the semi-supervised GW model (in orange) outperforms the fully supervised GW model (red), i.e. it reaches a similar contrastive loss value with fewer matched examples. This advantage is visible for both vision--proto-language alignment and (albeit to a lower extent) vision--natural language alignment.
    
    In summary, our main conclusion is also independently replicated on the \textit{Factory} dataset. Semi-supervision primarily helps in optimizing the translation property with fewer annotated bimodal data, while the inclusion of a GW in the architecture is important to ensure the alignment of multimodal representations. Overall, the model that best satisfies our two primary properties~\ref{prop:tr} and~\ref{prop:cont} is the one combining a GW architecture with a semi-supervised training setting (\ptrcontdcycy{}).

    For completeness, we also studied the influence of the total number of available unimodal samples ($M$) used for semi-supervised training (for the vision--proto-language case, as done previously). We fixed the number of bimodal matched training samples to $N=200$, as it was where the effect of semi-supervision was highest (gray line in Figure \ref{fig:properties_results}). 
    
    Figure \ref{fig:unaligned_examples_influence}B depicts very similar results as for the \textit{Simple Shapes} dataset. When enough additional unimodal unmatched data is available (roughly between $M+N = 1000$ and $M+N = 10,000$), the performance of semi-supervised models quickly surpasses the supervised ones. This implies that a dataset of $\sim10,000$ unpaired samples could have been sufficient to observe qualitatively similar behavior in our semi-supervised models as the full \textit{Factory} dataset with 200,000 samples. 
\end{changes}

\subsection{Secondary properties}

In addition to the \emph{primary} properties, we described in section~\ref{section:definitions} two desirable \emph{secondary} properties~\ref{prop:cy} and~\ref{prop:dcy} that multimodal networks could possess.
Figure~\ref{fig:properties_results} (right) shows the performance of the models on these secondary properties. Unsurprisingly, the models that explicitly optimize the corresponding losses reach the best performance: \ptrcy{} and \ptrcontdcycy{} for the cycle loss; \ptrdcy{} and \ptrcontdcycy{} for the demi-cycle loss. However, some of the other models also perform relatively well, even though none of them were explicitly trained for the secondary properties. We surmise that this could be due to the relations $R_1$-$R_2$ that exist between primary and secondary properties.
In line with relation $R_1$, we see that all models, as they are trained with a translation loss, improve their cycle loss when the number of annotations $N$ becomes sufficient. 
According to relation $R_2$, we see that the \ptrcont{} curve has a much lower demi-cycle loss than \ptr{} (in blue), and \ptrcontdcycy{} (in green) curves.

Here again, we can conclude that the best model to jointly satisfy the two \emph{secondary} properties is the one combining a GW architecture with self-supervised training (\ptrcontdcycy{}).

\subsection{Downstream tasks}
We have seen that a GW can help a multimodal system learn useful representations for translation and alignment. Logically, these improved multimodal representations should also facilitate performance in downstream tasks, in particular for multimodal transfer. We explicitly test this prediction by comparing the performance of our different trained models (with/without a GW) on two downstream tasks: the ``odd-one-out'' (OOO) and the shape classification tasks.

\subsubsection{Odd-one-out}
\begin{figure*}[htb!]
    \centering
    \includegraphics[width=0.9\linewidth]{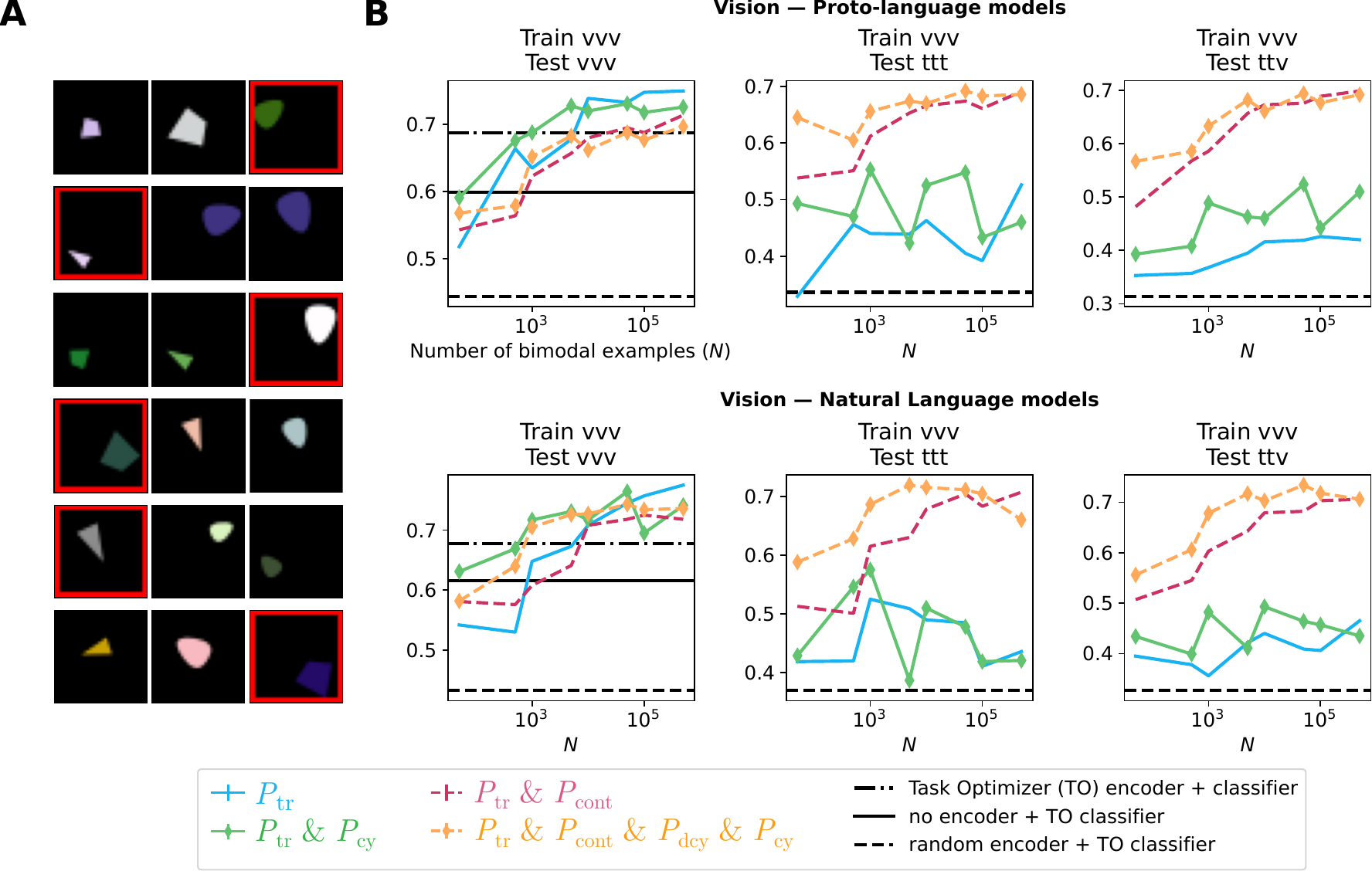}
    \caption{Downstream task. Panel A shows examples of the odd-one out (OOO) dataset. Each row correspond to a triplet of images where two images have at least one common attribute, and the third one (with the red border) differs from the other two. On panel B, we plot the results obtained on this task. Each point in the colored curves represents a model pre-trained on the domain translation and alignment task (using $N$ annotated training pairs, x-axis), with the encoders subsequently frozen. For each model, we train a new classifier to predict the odd-one out from the concatenation of three encoded visual representations ``\texttt{vvv}'' ($e_v(x_i)$). As baselines, we also train a new ``Task-optimized encoder + classifier'' end-to-end on the OOO task (dash-dotted line); we train a classifier directly on the visual latent representation instead of the encoded one (no encoder; solid black line); or we train a classifier based on an encoder with randomized weights (dashed line). In the first column, we show the results when testing the model on the trained domain ``\texttt{vvv}''. In the second column, we evaluate the ``\texttt{vvv}''-trained classifier on three language (or proto-language) representations ``\texttt{ttt}'' ($e_t(y_i)$). In the last column, we test it on a cross-modal version comprising two linguistic representations and one visual representation ``\texttt{ttv}''.}
    \label{fig:downstream-task}
\end{figure*}

For each trial of this task, three samples are given (at first, only in the visual domain). Two of them share at least one attribute (shape, size, color, position or orientation), while the last image differs from the other two across all attributes, and is thus considered the odd-one-out. Figure~\ref{fig:downstream-task} -- panel A, shows some examples: in the first row, the first two images share the same shape and a similar orientation, the odd-one-out is the third image.
Appendix~\ref{appendix:ooo-dataset} details how we construct the dataset, and how the triplets are selected.

To solve this task, we start by using the pretrained models from the previous sections, with their encoders frozen. We concatenate 3 encoded latent vectors $[e_v(x_1), e_v(x_2), e_v(x_3)]$. We then classify the OOO with a 2-layer classifier (from $3 \times 12$ to 16 hidden neurons, then 16 to 3 neurons) trained specifically on this task.
We additionally evaluate 3 baseline models:
\begin{enumerate}
    \item \textit{task-optimized (TO) encoder + classifier} where we keep the pretrained unimodal visual VAE to encode the image, then jointly train the encoder $e_v$ (from scratch) and classifier end-to-end for the downstream OOO task;
    \item \textit{no encoder + TO classifier} where we remove the encoder $e_v$ and directly learn the OOO classifier from the unimodal pretrained model;
    \item \textit{random encoder + TO classifier} where we use a random and frozen encoder $e_v$, and only train the classifier.
\end{enumerate}

Note that the first baseline has the same overall number of parameters than the evaluated models, but more trainable parameters (in the encoder $e_v$); the second baseline has the same number of trainable parameters, but fewer overall parameters; the last baseline has the same number of both overall and trainable parameters.

The first column in figure~\ref{fig:downstream-task}B plots the OOO accuracy obtained by the models when tested in the same condition as during training (\texttt{vvv}, i.e. comparing three images). All of our pretrained models (trained using translation and/or alignment objectives) outperform the three baselines when given enough supervision ($N\gtrapprox10,000$). This is even true for the strongest baseline trained end-to-end (``TO encoder + classifier''), which has the same architecture but more trainable parameters allocated for the downstream task. This shows that pretraining multimodal representations can be helpful for downstream visual tasks; however, the various pretrained models remain  qualitatively comparable in this setting. 

Then, we investigate whether the models trained with images (\texttt{vvv} condition) can generalize across domains (i.e., transfer learning): what would happen if we test these models using representations coming from language instead ($e_t(y)$)?
The second and third columns on panel B plot the models' transfer learning performance, respectively, in the \texttt{ttt} condition (new modality) and in a \texttt{ttv} condition (a truly cross-modal setting, where two of the representations come from language descriptions, and one from a visual one). In both cases, we see that only the models that have a Global Workspace (\ptrcont{} and \ptrcontdcycy{}) are able to properly generalize to the other domain; the models without a GW (``translation'' and ``trans. + full cycles''), on the other hand, suffer a considerable drop in performance when transferring across domains. Nonetheless, their performance remains above chance level (1/3rd), indicating that they can also transfer some (limited) knowledge across domains. Our hypothesis is that these models may have learned to solve the task by comparing the distances between encoded representations ($e_v(x)$), and selecting the furthest one as the odd-one out. This strategy could generalize, to some extent, to latent vectors from a different encoder ($e_t(y)$), even if the resulting representations are not actually aligned.  
In conclusion, we note again that models trained using a GW (and in particular, the one trained with semi-supervised learning, ``all sup.+all cycles'') outperform the other ones in a downstream task requiring domain transfer.

\begin{changes}
    \subsubsection{Shape Classification}
    As a second downstream task, we evaluate our models on a shape classification task on the Simple Shapes dataset (see Table~\ref{tab:category_cls_simple_shapes}). 
    We test two different settings: ``linear probe'' where we train a linear shape classifier from the multimodal (global workspace) representation, and ``zero-shot'' where we use the alignment property of the model in the GW to classify images by matching them with captions \textit{à la CLIP}.
    
    More specifically, for the linear probe setting, we train a linear classifier on 500,000 pairs of $(GW_v, s)$ or $(GW_t, s)$ where $GW_m$ is the global workspace representation of modality $m\in\{v,t\}$ and $s$ is the shape category of the object (i.e.\@ diamond, egg, triangle); we measure performance on an independent test set of 1000 samples.

    For the zero-shot setting, we generate 100 objects from each shape category with random attributes (rotation, size, position and color) and we create the associated 100 sentences with our heuristic. We keep the same 100 sentences for each category, and only change the shape information. For each shape, the sentences are encoded into the language encoder (BERT + projection).
    The class representatives (or ``prototypes'') are obtained either by averaging the 100 outputs of the language encoder and then encoding the average into the GW representation (column $GW_t(\overline{\text{BERT}})$), or by first encoding each sentence into the GW representation, and then averaging (column $\overline{GW_t(\text{BERT})}$). We then follow the ``zero-shot classification'' procedure of CLIP~\cite{radford_learning_2021} to match each image input to the most similar prototype.
    
    We see in the results presented in Table~\ref{tab:category_cls_simple_shapes} that the worst model is \pcont{}, trained only for alignment with contrastive learning (as done in the CLIP study). This model aligns the representations, but does not have a decoder, i.e. no broadcast (see Table~\ref{table:selected-models}).
    \ptrcont{} performs globally better than \pcont{}, due to its training with an additional translation loss that constrains the decoders and provides broadcast abilities.
    Finally, our ``target'' model \ptrcontdcycy{} performs best in all settings; this can be attributed to its improved GW, with the demi-cycle loss reinforcing alignment (Relation~$R_3$), and the cycle loss improving the broadcast ability (as demonstrated already by improved translation in Figure~\ref{fig:properties_results}).

\end{changes}

\begin{table}[t]
    \begin{blockChanges}
        \resizebox{\linewidth}{!}{
            \centering
            \begin{tabular}{@{}lllll@{}}
                \toprule
                {}              & \multicolumn{2}{c}{Linear Probe} & \multicolumn{2}{c}{Zero-shot}                                                                   \\
                \cmidrule(lr){2-3} \cmidrule(lr){4-5}
                {}              & $GW_v$                           & $GW_t$                        & $GW_t(\overline{\text{BERT}})$ & $\overline{GW_t(\text{BERT})}$ \\
                \midrule
                \ptrcontdcycy{} & \textbf{0.9989}                  & \textbf{0.9997}               & \textbf{0.8477}                & \textbf{0.9863}                \\
                \ptrcont{}      & 0.9901                           & 0.9921                        & 0.7402                         & 0.9746                         \\
                \pcont{}        & 0.8957                           & 0.9176                        & 0.5947                         & 0.8721                         \\
                \bottomrule
            \end{tabular}
        }
        \caption{\label{tab:category_cls_simple_shapes} Linear probe and zero-shot performance on three-way shape classification on the Simple Shapes dataset. All models were trained using all available paired data ($N=500,000$). For zero-shot classification, image inputs were encoded into the multimodal (GW) representation and compared with prototype vectors, calculated either by encoding the average BERT embedding of 100 class-representative text captions ($GW_t(\overline{\text{BERT}})$), or by averaging the encoded representations across captions ($\overline{GW_t(\text{BERT})}$).}
    \end{blockChanges}
\end{table}

\subsection{Effect of unpaired data}
Until now, we always used all available data as our unsupervised training sets $\mathcal{U}_v$ and $\mathcal{U}_t$, and only varied the number $N$ of paired multimodal examples in the supervised training set $\mathcal{S}$. Here, we analyze how the performance of our semi-supervised models depends on the size of the unsupervised training sets.
Let us define as $M$ the number of strictly unpaired examples in the dataset, such that $N + M$ is the total number of examples in the dataset. In our previous experiments, $M+N$ was always fixed at $M+N=500,000$ and $N$ varied from $0$ to $500,000$.

\begin{figure}
    \centering
    \includegraphics[width=\linewidth]{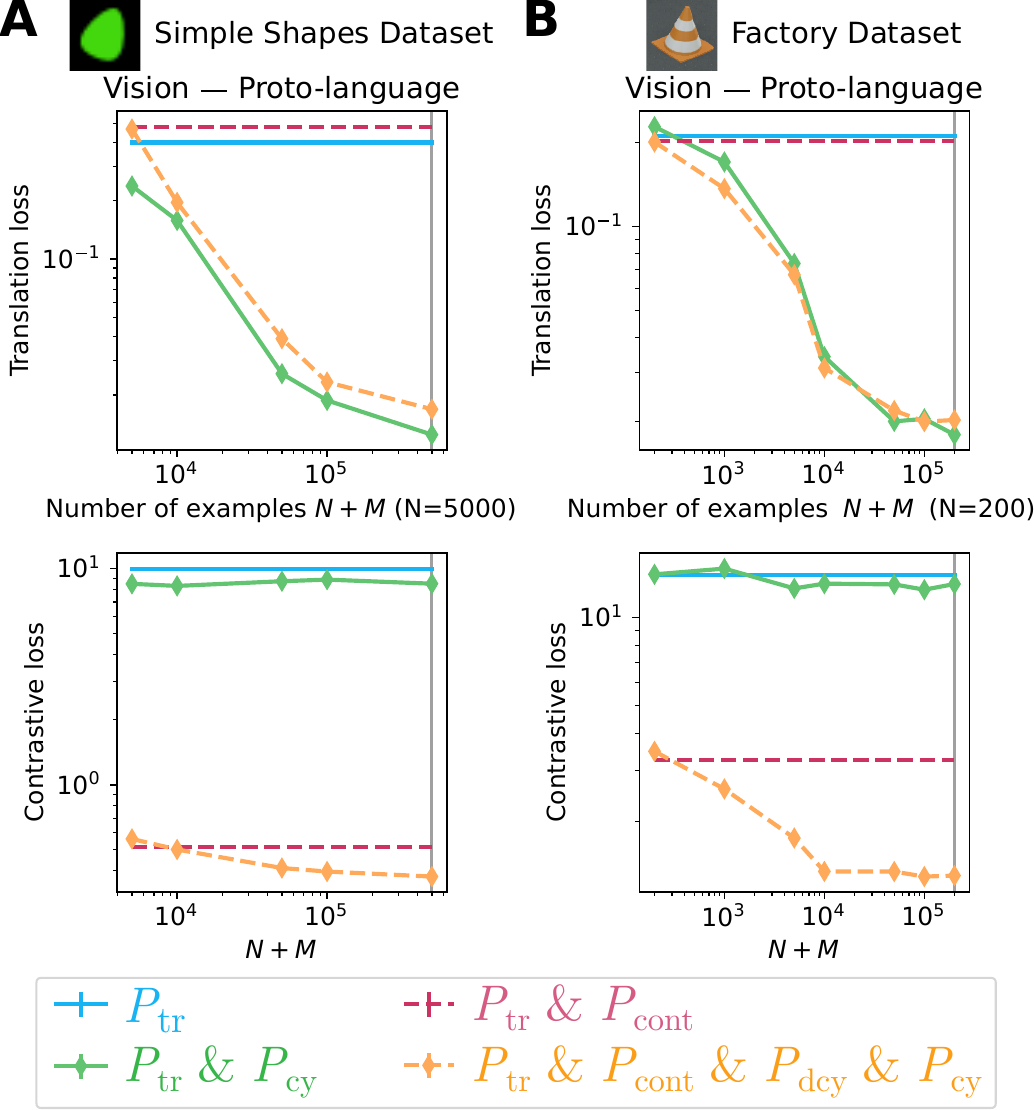}
    \caption{Influence of the number of unpaired examples $M$ on the models' performance. In order to facilitate comparisons, we increase $M$ by adding unimodal data samples to the same set, such that two models trained with $M1<M2$ unpaired examples will share the same $M1$ training examples. A) On the \textit{Simple Shapes} dataset, we fix $N=5000$, and $N+M$ varies from $5000$ to $500,000$. The results along the gray vertical line correspond to the results of figure~\ref{fig:properties_results}, first column, on the same gray line. B) The same measured is applied to the \textit{Factory} dataset. Here, we fix $N=200$ and $N+M$ varies from $200$ to $200,000$. The results along the gray vertical line correspond to the results of figure~\ref{fig:properties_results}, first column, on the same gray line. }
    \label{fig:unaligned_examples_influence}
\end{figure}

Figure~\ref{fig:unaligned_examples_influence}A shows the performance of new models trained with a fixed number $N=5,000$ of paired samples, and with $M$ increasing from $0$ to $495,000$ (so $N+M$ varies between $5,000$ and $500,000$).
That is, the results plotted in fig~\ref{fig:unaligned_examples_influence}A and those plotted in figure~\ref{fig:properties_results} (left column) can be envisioned as reflecting the same underlying ``3D surface'' where losses are expressed as a function of $N$ on one axis, and as a function of $N+M$ on the other; the two figures depict cross-sections of this same surface along orthogonal dimensions, intersecting on the gray vertical lines visible in each figure, i.e. $N=5,000$ and $N+M=500,000$. Indeed, the results along the gray lines are identical in both  plots.

The results reveal how the semi-supervised models make use of additional unpaired data. The purely supervised models (``translation'' and ``trans. + cont.'') only rely on the number of paired examples $N$ (resulting in horizontal lines in this figure). However, the semi-supervised models improve with additional unpaired samples on the translation and contrastive objectives. The figure also highlights that these improvements eventually saturate with increasing $M$. In other words, a dataset of roughly 50,000 unpaired samples could have been sufficient to observe qualitatively similar behavior in our semi-supervised models as the full \textit{Simple Shapes} dataset with 500,000 samples. 

%% file: sections/7_retrieval.tex
\section{Retrieval on COCO Captions}
\label{sec:retrieval}

\begin{changes}
    Until now, we have exclusively tested our models on synthetic data. It has allowed us to systematically test the effect of each loss on a controlled dataset (Simple Shapes and Factory). To verify the effectiveness of our method on a more natural dataset, we now train GW on the COCO Captions dataset.

    In the Simple Shapes and Factory experiments, all our unimodal modules were (variational) auto-encoders. This choice was made to allow us to visualize (as an image) the outcome of translations, cycles and demi-cycles, and help us evaluate the relevance of each model.
	This is unfortunately not as easily achievable with more natural datasets (auto-encoding models exist for natural images but are either too inaccurate, or too computationally intensive for our purposes). We thus decided to remove the auto-encoder constraint for this experiment, and used another proxy (than image reconstruction) to evaluate performance.
	We selected the COCO dataset~\cite{lin_microsoft_2014} for the experiments, and image/caption retrieval as the proxy task.
    
    To emphasize the importance of semi-supervised training, we only keep 20\% of the train\ +\ restval split as paired examples (\ptr{} and \pcont{}), and use the full training set for the unsupervised losses ($P_{cy}$ and $P_{dcy}$).
    We use a ResNet50~\cite{he_deep_2015} pretrained on ImageNet~\cite{deng_imagenet_2009} for our visual domain encoder, and BGE~\cite{xiao_c-pack_2023} for the text encoder.
    The full GW model (\ptrcontdcycy{}) is trained on the COCO dataset with the translation, demi-cycle, cycle and contrastive objectives, and compared against \pcont{} (a baseline trained only with a contrastive loss as in the CLIP study~\cite{radford_learning_2021}). We evaluate their retrieval performance as a downstream task without additional training.
    
    Table~\ref{tab:coco_retrieval} reports the average recall of the top K samples (R@K) on a 5-fold 1K random split of Karpathy's test split. We also provide the median rank of the correct sample among the 1,000 alternatives.
    Note that for caption retrieval there are 5 correct captions for each image and we keep the minimum rank of the 5 captions.
    We see that \ptrcontdcycy{} performs better than \pcont{}.
    These results go in the same direction as our previous results on the Simple Shapes and Factory datasets.
    They show that models trained with a semi-supervised GW have a better alignment, and better generalize to downstream tasks (here, retrieval). Note again that our models were never trained on the retrieval task, and the performance is only a consequence of the translation, contrastive, and cycle-consistency training.

   In addition to image-text retrieval, we can also evaluate the models trained on COCO for out-of-domain classification accuracy on ImageNet. This time, we use all training dataset examples for both supervised and semi-supervised objectives. To test the out-of-domain generalization, we learn a linear classifier from the visual GW representation to the 1000 classes of the ILSVRC~2012 ImageNet dataset. 
   A model trained solely with the contrastive loss (\pcont{}) classifies with  $50.3 \pm 0.2 \%$ accuracy and lags behind the full GW model trained with all losses (\ptrcontdcycy{})  at $55.7 \pm 0.2 \%$ accuracy (standard error of the mean across the 1000 classes of the accuracy difference between the two compared models). These results again show the advantage of our GW paradigm over a pure contrastive loss as used e.g. in the CLIP~\cite{radford_learning_2021} study.
\end{changes}

\begin{table*}[t]
    \begin{blockChanges}
        \centering
        \resizebox{\linewidth}{!}{
            \begin{tabular}{@{}lccccccccc@{}}
                \toprule
                \multirow{2}{*}{Model} & \multicolumn{4}{c}{Caption retrieval (i2t)} & \multicolumn{4}{c}{Image retrieval (t2i)} \\
                \cmidrule(lr){2-5} \cmidrule(lr){6-9}
                {}                     &  R@1                              &  R@5                             &  R@10       &  med rank     &  R@1        &  R@5        &  R@10       &  med rank \\
                \midrule
                \ptrcontdcycy{}        & $34.4 \pm 0.6$                       & $65.1 \pm 0.3$                      & $79.0 \pm 0.2$ & $2.8 \pm 0.2$ & $30.3 \pm 0.5$ & $65.3 \pm 0.1$ & $79.0 \pm 0.2$ & $3.0 \pm 0.0$ \\
                \pcont{}               & $31.8 \pm 0.6$                       & $62.5 \pm 0.5$                      & $76.0 \pm 0.4$ & $3.0 \pm 0.0$ & $27.3 \pm 0.2$ & $62.0 \pm 0.4$ & $77.3 \pm 0.3$ & $3.6 \pm 0.2$ \\
                \bottomrule
            \end{tabular}
            }
        \caption{\label{tab:coco_retrieval} Image-Text retrieval performance for models trained with 20\% paired data from COCO train set + restval of the Karpathy~\cite{karpathy_deep_2015} 1K split (averaged over 5 runs).}
    \end{blockChanges}
\end{table*}

%% file: sections/8_discussion.tex
\section{Discussion}

\subsection{Summary}
The Global Workspace theory offers an account of multimodal integration in the human brain~\cite{baars_cognitive_1993, baars_global_2005, dehaene_neuronal_1998}. Prior work  \cite{vanrullen_deep_2021} has provided theoretical insights on how to use a GW architecture to connect the latent spaces of pretrained deep neural networks for different modalities. In this work, we present an initial empirical validation of some of these ideas, by investigating the bimodal (vision-language) integration abilities that emerge with vs. without a GW latent space. We further explore the possibility of improving the training via unsupervised objectives, by varying the amount of matched (bimodal) data available for training the GW encoders and decoders.

Our results show that semi-supervision is particularly important to achieve efficient vision-language \textit{translation}. Semi-supervised models---with unsupervised cycle-consistency losses---needed approximately 4-7 times fewer annotated bimodal examples than their fully-supervised counterparts to reach the same bimodal translation accuracy. 
The GW architecture, on the other hand, is critical for \textit{(contrastive) alignment} between the vision and language representations. 
Overall, the semi-supervised GW model proved the best at jointly satisfying the two primary properties that we advocate for multimodal systems: \textit{translation} \& \textit{contrastive alignment}.
Furthermore, we showed that our semi-supervised GW pretraining method produced meaningful and better multimodal representations for downstream tasks than a dedicated model with the same number of weights. These aligned multimodal GW representations allowed the system to transfer knowledge from one modality to the other (i.e., bimodal and cross-modal domain transfer).

\subsection{Limitations and open questions}
Due to limited computational resources, all reported experiments were conducted with only one repetition (with an unoptimized seed set to 0 from the beginning). Thus, repeating all experiments with different random seeds for weight initialization and/or dataset splits (e.g. for the selected subset of $N$ matched exemplars) could yield smoother performance curves and allow us to estimate statistical variability. However, we do not expect this to affect our general conclusions.

Similarly, we had to limit the present study to two handcrafted and relatively simple bimodal datasets. Despite their simplicity, one advantage of these datasets was our ability to quickly and parametrically control image and text generation properties, which facilitated our understanding of the various model components. The fact that our main conclusions could be replicated across these two datasets already hints at their generality. However, an important next step would be to extend the study to more realistic, large-scale bimodal datasets and benchmarks. 
Paradoxically, the inherent diversity and richness of real-world data, instead of impeding our system, could result in a more precise bimodal alignment from unimodal data, and thus lead to better semi-supervised training and improved generalization---provided that sufficient computational resources would be available to train our models on such large-scale datasets.

A related question is whether our findings could generalize to bimodal translation and alignment problems when the two domains are not bijectively related. For instance, images and proto-language descriptions in our datasets were bijectively related, in the sense that a unique attribute vector could be inferred from each image, and vice-versa. Comparing this situation to the vision/natural language setting (where linguistic ambiguities, synonymy, categorical terms, etc. challenged the one-to-one mapping between domains), we already saw that our approach can still work without a perfect bijection. However, it worked less well than when using proto-language. Would it still work at all if the multimodal correspondence was only very loosely defined, e.g. matching impressionnist paintings to Baudelaire poetry? If not, could the presence of additional modalities (e.g. a limbic system encoding emotions; an auditory system to process rhymes and prosody) help the model resolve ambiguities? These are exciting questions for follow-up studies.

In addition to the multimodal integration abilities that we already demonstrated here, additional properties could be expected from our GW model that could be tested in future studies. 
In particular, prior work~\cite{devillers_does_2021} has shown that models trained only with a contrastive loss to align information across domains (such as CLIP~\cite{radford_learning_2021}) tend to filter out domain-specific data. This side-effect hinders their unimodal generalization performance, e.g. when comparing against expert visual models~\cite{devillers_does_2021}. 
Our setting combining translation, contrastive alignment and semi-supervised cycle-consistency objectives could be a way around this problem. Specifically, the demi-cycle property should help align multimodal representations, while at the same time forcing the encoder to retain domain-specific information. This should result in preserved unimodal generalization abilities, compared to models trained only for contrastive alignment. In other words, we could expect a GW model to learn to represent the union of the two domains, rather than just their intersection.


\subsection{Future model extensions}
\label{future}
The proposed strategy proved successful for bimodal vision-language integration. However, much remains to be incorporated into our model's architecture before this promising approach could be considered a full implementation of the GW theory, and thereby possibly rival with human-level capabilities. 

A first extension would be to increase the number of domains and modalities integrated into the workspace. A simple way to  achieve this within our current architecture could be to use $n$ specialist-modules instead of just 2, yet only one module would access the GW at any given time. Training such a system could still be performed via a combination of unsupervised cycle-consistency objectives on unimodal data, and supervised training from pairwise matched data, as done in the present study. Such an extension would resemble the recently proposed extension of the CLIP vision-language alignment model~\cite{radford_learning_2021} to the new ``ImageBind'' model, aligning several distinct modalities to a visual representation space~\cite{girdhar_imagebind_2023}. Our version, however, would be centered around a GW latent space, with a symmetric architecture that does not favor vision compared to other modalities, and using the necessary encoders and decoders to satisfy both translation and cycle-consistency properties. 

In a second step, given the availability of multiple modules and their encoders and decoders to/from the GW, it could become useful to allow two or more modules to simultaneously encode their representations in the GW. As in the original formulation of the theory by Baars~\cite{baars_cognitive_1993,baars_global_2005}, this would require a dedicated attentional system to control access to the GW, and some sort of attention-dependent fusion mechanism to combine information from these modalities into the GW latent space.

Finally, another possible extension may be to include recurrent dynamics into the model. Having a model that can maintain its internal state over time, but can also update it based on novel information from the internal or external environment, is useful in many domains (planning, robotics, ...). Moreover, more advanced modules could be envisioned in this dynamic context, such as a memory module~\cite{graves_neural_2014,graves_hybrid_2016} or a world model~\cite{hafner_mastering_2023}.

\subsection{Implications for Cognitive Science}
The GW is a prominent theory of higher brain function. 
The present results already show that it is possible to start implementing this sort of cognitive strategy in multimodal deep learning systems. In particular, the demonstrated feasibility of using semi-supervised learning techniques can go some way towards reconciling these models with human multimodal learning (which relies on much less explicit supervision than the standard deep learning approaches).
Nonetheless, the present findings by themselves do not prove or disprove the GW theory, as they do not yet constitute a full implementation of the GW framework~\cite{vanrullen_deep_2021}. However, with the extensions proposed above (section~\ref{future}: additional modalities, attention control system, recurrent implementation with temporally extended inputs and outputs), it might become possible, in the relatively short term, to use this sort of artificial system to draw conclusions of relevance to Cognitive Science.

%% file: appendices/code.tex
\section{Code availability}
We provide the code for our experiments, pretrained models, and our datasets, here: \url{https://github.com/bdvllrs/bimGW}.

%% file: appendices/architectures.tex
\section{Architecture details}
\label{appendix:architecture}

We describe here the architecture and training details of the specialist modules for vision and natural language (and for the two datasets). These modules were subsequently frozen before connecting them to the GW as described in the main text.

\subsection{Visual VAE for Simple Shapes dataset}
The VAE architecture is inspired by the model used in \cite{ghosh_variational_2020} for $32\times32$ images.

In the encoder, all convolutions have a padding of 1, a stride of 2, and a kernel-size of 4.
In the decoder, the transposed convolutions have a padding of 1, a stride of 2, and a kernel size of 4, except the first one which has a stride of 1. The final convolution has a stride of 1 and a kernel size of 4.

\begin{center}
\begin{tabular}{l|l}
Encoder & Decoder \\
\hline
$x\in \mathbb{R}^{3\times 32\times 32}$ & $z\in \mathbb{R}^{12}$ \\
$\text{Conv}_{128} - \text{BN} - \text{ReLU}$ & $\text{FC}_{8\times8\times 1024}$ \\
$\text{Conv}_{256} - \text{BN} - \text{ReLU}$ & $\text{ConvT}_{512}-\text{BN}-\text{ReLU}$ \\
$\text{Conv}_{512} - \text{BN} - \text{ReLU}$ & $\text{ConvT}_{256}-\text{BN}-\text{ReLU}$ \\
$\text{Conv}_{1024} - \text{BN} - \text{ReLU}$ & $\text{ConvT}_{128}-\text{BN}-\text{ReLU}$ \\
$\text{Flatten} - \text{FC}_{2\times 12}$ & $\text{Conv}_{1}-\text{Sigmoid}$ \\
\end{tabular}
\end{center}

\subsection{Text VAE for Simple Shapes dataset}
The specialist text model (for the natural language modality) preprocesses the English sentences using the pretrained BERT model~\cite{devlin_bert_2019}, followed by a trained VAE (with an additional classifier loss) to reduce the domain dimensionality from 768-dim (BERT output) to 12 latent dimensions.
The module comprises an encoder-decoder model, and several classifier heads predicting (i) the image attributes, and (ii) the details of the sentence grammar. The sentence grammar represents all the choices made by the text-domain heuristic to generate a sentence (word orders, word variants, and synonyms; see Appendix~\ref{appendix:generating-text}). We use one separate classifier head to predict each of the independent choices made by the text-generation heuristic (see~\ref{appendix:generating-text}) to generate the sentence\footnote{There are 39 different grammar classifier heads: one predicts in which order the attributes will appear in the sentence (18 classes), another one predicts the beginning of the sentence, others predict variations in the wording.}. 
The text-domain VAE encoder uses the SymLog activation function from \cite{hafner_mastering_2023}. The final ``specialist module'' for natural language inputs uses the following architecture.

\begin{center}
\begin{tabular}{l|l|l}
Encoder & Decoder & Attr. Classifier \\
\hline
$x\in \mathbb{R}^{768}$ & $z\in \mathbb{R}^{12}$  & $z \in \mathbb{R}^{12}$ \\
$\text{FC}_{768} -\text{ReLU}$ & $\text{FC}_{384} -\text{ReLU}$ & $\text{FC}_{12} -\text{ReLU}$ \\
$\text{FC}_{384} -\text{ReLU}$ & $\text{FC}_{768} -\text{ReLU}$ & $\text{FC}_{11}$\\
$\text{FC}_{2\times 12} -\text{SymLog}$ & $\text{FC}_{768}$ & \\
\end{tabular}
\end{center}

The grammar classifiers follow the same structure as the attribute classifier, with a number of output neurons in the final FC layer dependent on the number of classes.

The final loss is defined as:

\begin{equation}
    \mathcal{L}_\text{t} = \alpha_{VAE}\mathcal{L}_{VAE} + \alpha_{CLS}\left(\mathcal{L}_\text{attr.} + \frac{1}{G}\sum_{i=1}^{G}\mathcal{L}^i_{\text{grammar}}\right)
\end{equation}
where $G=39$ is the number of grammar classifiers.

We obtained the best results when giving a higher weight to the classification loss than to the VAE loss, i.e., $\alpha_{VAE}=1$ and $\alpha_{CLS}=1,000,000$.

\subsection{Visual VAE for Factory dataset}

\begin{figure}
\centering
\includegraphics[width=\linewidth]{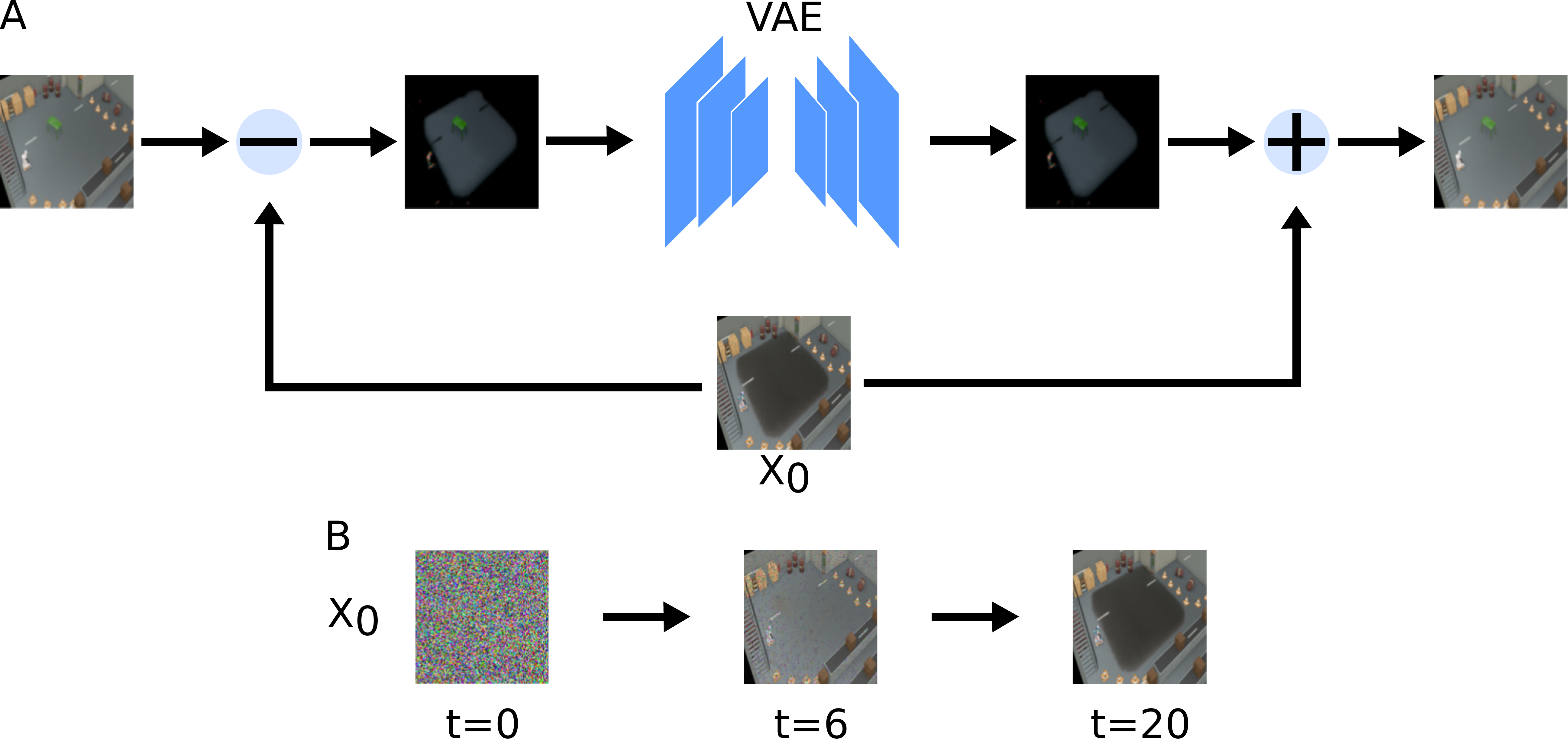}
\caption{A) Illustration of the architecture of the adapted VAE. Before passing the image $X$ to the encoder, we subtract from it a learnable tensor $X_0$, which is then added back to the decoder output before computing the VAE reconstruction loss. The goal of this tensor is to capture the fixed background such that only the changing image elements (i.e., the table and its immediate suroundings) are passed to the VAE. B) Evolution of $X_0$ through successive training epochs $t$.}
\label{fig:Appendix:factory:VAE}
\end{figure}

To facilitate disentangling the fixed from the changing image elements, we modified the standard VAE architecture with a learnable image tensor $X_0$ having the same size as the images of the \textit{Factory} dataset, $X_0 \in \mathbb{R}^{3\times 128\times 128}$. It was initialized randomly with a uniform distribution between 0 and 1, and trained jointly with the VAE that learned to reconstruct the image difference $X-X_0$. Figure \ref{fig:Appendix:factory:VAE}B illustrates how $X_0$ learned to capture the fixed background elements. We can see that the image locations potentially associated with the table are darker than the rest; this makes it possible to highlight the table in the subtracted image $X-X_0$, particularly when the table is dark. 


In the encoder, convolution layers have a kernel size of 3, a stride equal to 2 and a padding of 1.
For the decoder, the first transposed convolution layer has a kernel size of 7, a stride of 8 and an output padding of 1. All the other transposed convolution layers have a kernel size of 4, a stride of 2 and a padding equal to 1.

\begin{center}
\begin{tabular}{l|l}
Encoder & Decoder\\
\hline
$x\in \mathbb{R}^{3\times 128\times 128}$ & $z\in \mathbb{R}^{10}$\\
$\text{Conv}_{64} - \text{BN} - \text{ReLU}$ & $\text{FC}_{8\times8\times 512}$ \\
$\text{Conv}_{128} - \text{BN} - \text{ReLU}$ & $\text{ConvT}_{256}-\text{BN}-\text{ReLU}$ \\
$\text{Conv}_{256} - \text{BN} - \text{ReLU}$ & $\text{ConvT}_{128}-\text{BN}-\text{ReLU}$ \\
$\text{Conv}_{512} - \text{BN} - \text{ReLU}$ & $\text{ConvT}_{64}-\text{BN}-\text{ReLU}$ \\
$\text{Flatten} - \text{FC}_{2\times 10}$ & $\text{Conv}_{1}-\text{Sigmoid}$ \\
\end{tabular}
\end{center}

\subsection{Text VAE for Factory dataset}

As for the \textit{Simple Shapes} dataset, the specialist text module (for the natural language domain) is trained as a VAE with an additional classification loss. Here, we found best results using $\alpha_{VAE}=1$ and $\alpha_{CLS}=100,000$. We also use the SymLog activation function for the text encoder.
As for \textit{Simple Shapes}, we use an independent classifier head to predict each choice made by the heuristic to generate the sentence. In this case, we have 17 different classifier heads: one for the sentence structure, 9 for variations in the sentence and 7 predicting which type of adjective is used for each attribute.

\begin{center}
\begin{tabular}{l|l|l}
Encoder & Decoder & Attr. Classifier \\
\hline
$x\in \mathbb{R}^{768}$ & $z\in \mathbb{R}^{20}$  & $z \in \mathbb{R}^{20}$ \\
$\text{FC}_{768} -\text{ReLU}$ & $\text{FC}_{384} -\text{ReLU}$ & $\text{FC}_{512} -\text{ReLU}$ \\
$\text{FC}_{384} -\text{ReLU}$ & $\text{FC}_{768} -\text{ReLU}$ & $\text{FC}_{512} -\text{ReLU}$\\
$\text{FC}_{2\times 20} -\text{SymLog}$ & $\text{FC}_{768}$ & $\text{FC}_{6}$\\\\
\end{tabular}
\end{center}

%% file: appendices/language_domain_generation.tex
\section{Text Generation Heuristics}
\label{appendix:generating-text}
\subsection{Simple Shapes Dataset}
The text domain describes the images in natural language (English) and is automatically generated from the attribute vectors. Our aim was to create varied descriptions of the images while still describing them in sufficient detail.
Before setting the text generation rules, we asked a number of human participants to describe a few sample images from our dataset ``in their own words but as accurately as possible, such that a faithful reproduction of the image could be obtained from their description''. We gathered 100 such descriptions, and then chose our text generation heuristics so as to produce similar descriptions.
The generator first transforms the attribute values into words, then assembles them into sentences.

For the shapes, we use several descriptions of the three shapes given by the participants (e.g. the diamond shape could be described as a ``kite'', a ``diamond'' or a ``arrow-shaped polygon'').
The rotation angle is described either using 16 cardinal points (``west'', ``northwest'', ``north-northwest'', ...), corner locations (``left'', ``top-left'', ``top top-left''), or an actual angle (``rotated 15 degrees clockwise'') where the angle in degrees is approximated by the closest multiple of 5.
The sizes are split between 4 categories (``tiny'', ``small'', ``medium'' and ``large''). For the location, the image is divided into a $7\times7$ grid and described accordingly (``at the very top, slightly left'', ``in the center'', ...). For the colors, we select 141 color names and the corresponding RGB values from matplotlib's \cite{Hunter:2007} list of named colors; the name of the closest color in RGB space is selected.

The final sentence is generated by sampling the attribute descriptions as explained above, and a predefined set of linking words or phrases between them (for example, ``There is an isosceles triangle, it's located top, slightly right, it's large, it's in dark slate grey colored, it's pointing towards the north-northwest.'', or ``The image represents an isosceles triangle, it's lower left, it's large, it's white smoke color, it's pointing to the top.'').

In total, the text generator makes 39 choices in order to create one sentence. These choices include the order of appearance of the attributes, linking words, or the choice of synonyms.

\subsection{Factory Dataset}

As for the \textit{Simple Shapes} dataset, we used heuristic rules to create natural language (English) descriptions of the images based on the table attributes. The orientation is described using different cardinal points (''North'', ''North-West'', ''West'', ...) or an absolute angle $\theta$ (with a precision of 5 degrees), with $\theta \in [0, 2\pi[$. The location of the table (which is originally in $([-4, 2], [-2, 4])$ in the simulated environment coordinates) is described either by a position in the image (''on top of the image'', ''in the center'', ''between the center and the left of the image'', ...) or by its relation to other objects in the scene (''close to the stairs'', ''near the left part of the conveyer belt'', ...). For the color, we split the Hue space ($Hue \in [0,1[$) in 15 classes (keeping Saturation and Lightness equal to one). Each class is associated with a color name. 
The final sentence is generated in same way as for the \textit{Simple Shapes} dataset, with attributes and linking words chosen randomly.

%% file: appendices/odd_one_out.tex
\section{odd-one out dataset}
\label{appendix:ooo-dataset}
To generate each triplet used in the odd-one-out downstream task, we first randomly select a reference image from our \textit{Simple Shapes} dataset. Then, we select a common attribute by uniformly sampling among ``shape'', ``location'', ``size'', ``orientation'' and ``color''. Next, we select the image (out of the entire dataset) that has the closest selected common attribute with the reference image; this will be the positive (matching) image (in case of multiple matches, we randomly select one of them). Finally, to select the negative (odd-one out) image, we define a comparison function between one image $x$, and two other images $x_1, x_2$ as $\min(d(x, x_1), d(x, x_2))$, where $d(x_i, x_j)$ is the minimum distance among all attributes\footnote{Let $\mathcal{A}$ be se set of attributes (``category'', ``position\_x'', ``position\_y'', ...), then $d(x_i, x_j) = \min_{a\in\mathcal{A}} |x_i^{(a)} - x_j^{(a)}|$.}. Using this comparison function, we sort all other images by their distance to the two selected images (reference and matching image), and randomly select one of the 500 furthest ones.

We generate a dataset of $500,000$ such training triplets, and $1000$ test triplets.